\documentclass{article}

\usepackage{arxiv}

\usepackage[utf8]{inputenc} % allow utf-8 input
\usepackage[T1]{fontenc}    % use 8-bit T1 fonts
\usepackage{hyperref}       % hyperlinks
\usepackage{url}            % simple URL typesetting
\usepackage{booktabs}       % professional-quality tables
\usepackage{amsfonts}       % blackboard math symbols
\usepackage{nicefrac}       % compact symbols for 1/2, etc.
\usepackage{microtype}      % microtypography
\usepackage{lipsum}
\usepackage{graphicx}
\usepackage{booktabs}
\usepackage{multirow,array}
\usepackage{threeparttable}
\usepackage{algorithmic}
\usepackage{graphicx}
\usepackage{longtable}
    \usepackage{array}
\usepackage{subcaption}
\usepackage{microtype}
\usepackage{adjustbox}
\graphicspath{ {./Fig/} }

\title{Applications of Generative Adversarial Networks in Anomaly Detection: A Systematic Literature Review}

\author{
 Mikael Sabuhi \\
  Electrical and Computer Engineering\\
  University of Alberta\\
  Edmonton, Canada \\
  \texttt{sabuhi@ualberta.ca} \\
   \And
 Ming (Chloe) Zhou \\
  Electrical and Computer Engineering\\
  University of Alberta\\
  Edmonton, Canada \\
  \texttt{mzhou4@ualberta.ca} \\
  \And
 Cor-Paul Bezemer \\
  Electrical and Computer Engineering\\
  University of Alberta\\
  Edmonton, Canada \\
  \texttt{bezemer@ualberta.ca} \\
    \And
 Petr Musilek \\
  Electrical and Computer Engineering\\
  University of Alberta\\
  Edmonton, Canada \\
  \texttt{pmusilek@ualberta.ca} 
}

\begin{document}
\maketitle
\begin{abstract}
Anomaly detection has become an indispensable tool for modern society, applied in a wide range of applications, from detecting fraudulent transactions to malignant brain tumours. 
Over time, many anomaly detection techniques have been introduced. However, in general, they all suffer from the same problem: a lack of data that represents anomalous behaviour. As anomalous behaviour is usually costly (or dangerous) for a system, it is difficult to gather enough data that represents such behaviour. This, in turn, makes it difficult to develop and evaluate anomaly detection techniques.
Recently, generative adversarial networks (GANs) have attracted a great deal of attention in anomaly detection research, due to their unique ability to generate new data. In this paper, we
present a systematic literature review of the applications of GANs in anomaly detection, covering 128 papers on the subject.
The goal of this review paper is to analyze and summarize: (1)~which anomaly detection techniques can benefit from certain types of GANs, and how, (2)~in which application domains GAN-assisted anomaly detection techniques have been applied, and (3)~which datasets and performance metrics have been used to evaluate these techniques. Our study helps researchers and practitioners to find the most suitable GAN-assisted anomaly detection technique for their application. In addition, we present a research roadmap for future studies in this area.
\end{abstract}

% keywords can be removed

\section{Introduction}\label{sec:Introduction}

% P1:
%In modern society, many systems depend on and generate enormous amounts of data. This data is important for many of the decision-making processes in these systems. Normally, a system operates under the expected conditions. However, in rare cases an anomaly may occur. 
% Such anomalies can have a disastrous impact on the system itself, or its environment. Therefore,  it is important to be able to detect such anomalies as early as possible to lower the impact. For example, *some example about early detection of tumours* + *example about preventing a riot through surveillance*

% P2: Because of the importance of anomaly detection, it has received widespread attention in research in the past years. Despite the amount of progress in this research area, there are still an important open challenge: acquiring data about anomalies that can be used to test anomaly detection techniques.

% P3: Explain why GANs are good candidates to help with this challenge.

In modern society, many systems depend on and generate enormous amounts of data. This data is important for many decision-making processes related to these systems. Normally, systems operate under the expected conditions. However, in rare cases, anomalies may occur. Such anomalies can have a disastrous impact on the system itself or on its environment. Therefore,  to lower the impact, it is important to be able to detect such anomalies as early as possible. 
For example, cancer is an anomaly in human tissue. Breast cancer is the second leading cause of cancer death in women~\cite{sun2017risk}. According to a recent study by the American Cancer Society~\cite{siegel_cancer_2020}, breast cancer alone accounts for 30\% of female cancers. Early detection and treatment of breast cancer would highly increase the chance of survival~\cite{ginsburg2020breast}. %Therefore, it is essential to develop techniques for detecting breast cancer in the early stages~\cite{ott2009importance}. 
Similarly, with an increasing need to ensure public safety in crowded areas, development of real-time video surveillance systems becomes unavoidable. It is critical to seamlessly monitor the crowd to immediately detect anomalous (or \emph{abnormal}) movements to help prevent theft~\cite{mandal2016automatic}, vandalism~\cite{ghazal2007real}, and terrorist attacks~\cite{mould2014video}.

The process of finding the anomalous behaviour of a system is referred to as \textit{anomaly detection}. The primary objective of anomaly detection is to differentiate between the expected and unexpected behaviour of a system.
Considering the importance of anomaly detection, it has received widespread attention in research. Despite the progress in this research area, there is still an important open challenge: the acquisition of data about anomalies that can be used to test anomaly detection techniques.

A recent trend in anomaly detection is the use of generative adversarial networks (GANs). Proposed by Ian Goodfellow et al.~\cite{GANGoodfellow} in 2014, GANs are a type of unsupervised generative model which gained much attention from the research community. A well-trained GAN can generate realistic-looking data by sampling from a learned data distribution. 
A GAN consists of a generator and a discriminator model. These two models are pitted against each other in a two-player zero-sum game situation, iteratively improving their capabilities to generate and discriminate data. %The generator's goal is to fool the discriminator in distinguishing the generated data from the original data by improving its generative capability. The discriminator's goal is to correctly identify the fake data from the original data by improving its discriminative capability. The zero-sum games are often solved by the minimax theorem, i.e., by minimizing the opponent's maximum profit. 

The ability of GANs to generate data makes them attractive for anomaly detection research from two perspectives. First, they can potentially help generate the hard-to-acquire anomalous data points. Second, they can be used to learn the distribution of the data for the normal operating condition of a system and act as an anomaly or outlier detector.

%P3: Explain why GANs are good candidates to help with this challenge.

In this paper, we conduct a systematic literature review of the applications of GANs for anomaly detection. We address the following research questions (RQs): 
\begin{itemize}
    \item \textbf{RQ1: What is the role of GANs in anomaly detection?}
We identified two roles that GANs play in anomaly detection: data augmentation and representation learning. In contrast to the remarkable ability of GANs to generate realistic-looking data, most of the reviewed papers use them for representation learning rather than data augmentation. The reason for this inclination is that, despite the improvement in the anomaly detection accuracy after data augmentation, the reported improvements are not substantial. When GANs are used for data augmentation in anomaly detection, we refer to it as GAN-assisted anomaly detection. The other role of GANs in anomaly detection is representation learning. In this case, the examined papers use the data from the normal class for training a GAN to learn the distribution of the normal data. A score is assigned to the new data by defining a score function, and the anomalous data in the test stage is identified based on a specific threshold. We refer to these techniques as GAN-based anomaly detection. 

    \item \textbf{RQ2: What are the application domains of anomaly detection with GANs?}
    The primary application areas where GANs are used for anomaly detection are medicine (19\%), surveillance (15\%) and intrusion detection (13\%).

    \item \textbf{RQ3: Which GAN architecture is used most often in anomaly detection systems?}
     We identified 21 architectures of GANs that are used for anomaly detection. Among these architectures, deep convolutional GANs (DCGANs) (32\%), standard GANs (23\%), and conditional GANs (16\%) are the most commonly used.

    \item \textbf{RQ4: Which type of data instance and datasets are most commonly used for anomaly detection with GANs?}
    50\% of the proposed GAN-based anomaly detection techniques use image datasets for anomaly detection purposes. Before being fed to the anomaly detection algorithms, the data are usually preprocessed. The most common preprocessing methods are resizing (23\%), normalization (19\%), and cropping (13\%).
    
    \item \textbf{RQ5: Which metrics are used to evaluate the performance of GANs in generating data and anomaly detection?}
    Only 21\% of the studied papers evaluated the GAN's performance in generating synthetic data, either in data augmentation or representation learning. Structural similarity indices (SSIM) (26\%) and peak signal-to-noise ratio (PSNR) (26\%) are the most commonly used metrics. Visual inspection to evaluate the quality of the generated data was reported in 5\% of the studied papers.
    To evaluate performance GANs in anomaly detection applications, 53\% of the primary studies used the area under the receiver operating characteristic curve (AUROC). 
    
    \item \textbf{RQ6: Which anomaly detection techniques are used along with GANs?}
    GAN-based anomaly detection is mostly done in a semi-supervised manner. DCGANs and standard GANs are the most popular architectures in semi-supervised anomaly detection using GANs. 
    % These GAN-based anomaly detection techniques are often compared to autoencoder-based methods.
    In supervised learning based anomaly detection, GANs are used to augment the dataset for the anomalous class. However, the studied papers report only minor improvements in the performance of anomaly detection techniques after augmenting the dataset with GANs. Only a few primary studies focused on pure unsupervised anomaly detection based on GANs, most using the standard version of GANs. Similar to semi-supervised techniques, unsupervised GAN-based anomaly detection techniques are mostly compared with autoencoder-based approaches. 
\end{itemize}

The findings presented in this survey will help researchers and practitioners to find the most suitable GAN-based anomaly detection techniques for their applications.

The rest of this paper is organized as follows. Section~\ref{sec:Background} provides a brief introduction to GANs. Section~\ref{sec:Method} describes the methodology used for conducting this systematic literature review. Section~\ref{sec:Results} presents the results of the review. Section~\ref{sec:futurework} discusses the open challenges and provides directions for future research. Section~\ref{sec:Threats} identifies the threats to validity of the review, and Section~\ref{sec:Conclusion} concludes the paper.

\section{Generative Adversarial Networks}\label{sec:Background}
%In this section, we give a brief introduction to GAN related backgrounds necessary to follow this systematic literature review. 

In 2014, Goodfellow et al.~\cite{GANGoodfellow} introduced a framework for estimating generative models based on an adversarial process.
This framework consists of two deep neural network-based models: a generative model $G$ and a discriminator model $D$. Model $G$ learns the training data distribution and uses it to generate new samples. Model $D$ determines whether a sample comes from the training data or was generated by the generative model. 
The power of GANs comes from the adversarial process, in which the two %deep neural network-based
models %($G$ and $D$)
are competing against each other to improve their accuracy in the designated task.\looseness=-1 

\begin{figure}[tbp]
  \centering
  \includegraphics[width=0.7\linewidth]{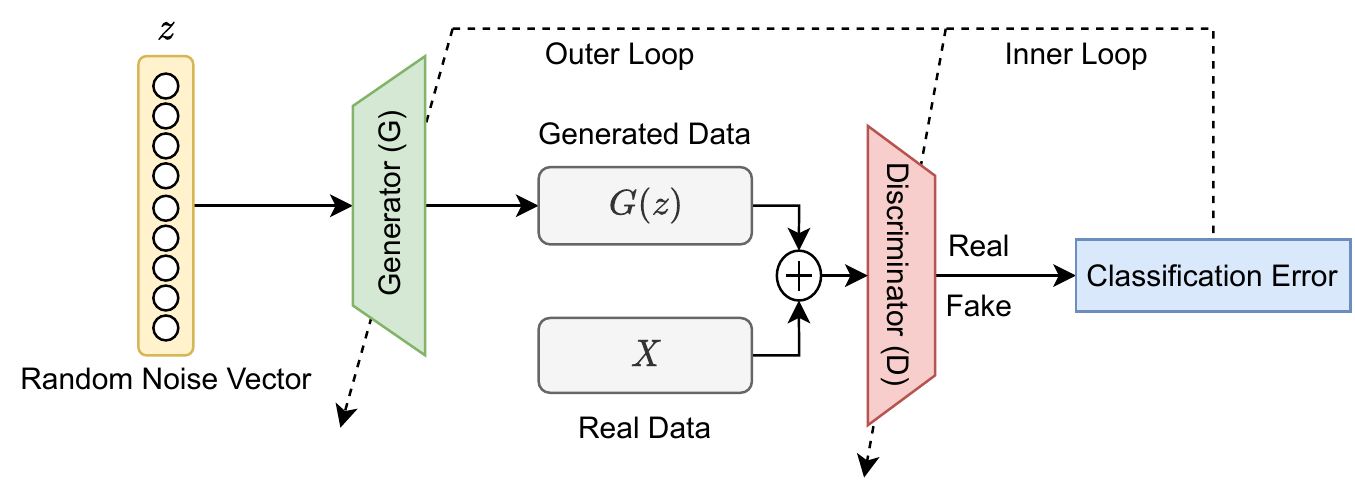}
  \caption{The building blocks of GANs. The classification error is used to update the parameters of the discriminator and generator models (shown by dashed lines). 
%   The dashed lines show the updating parameters of the crossing blocks.
  }
  \label{fig:buildingblocks}
  \end{figure}
  
  The diagram in Figure~\ref{fig:buildingblocks} shows the building blocks of a GAN~\cite{langr2019gans}:
  
  \begin{itemize}
    \item The \textit{Real Data (X)}, or the training dataset, contains the instances that the generator $G$ should learn to generate, usually in the form of a batch.
    \item \textit{Random Noise Vector ($z$)} is the raw input  to the generator. It is a vector of random numbers which the generator uses to generate fake examples.
  
    \item The \textit{Generator model (G)} is trained to learn the distribution of the input data. This model uses the input $(z)$ to generate fake examples ($G(z)$) that are indistinguishable from the real data.
  
    %\item \textit{Generated data G(z)} is a batch of data produced by the generator $G$. The dimensions of the generated data match those of the input data. 

    \item The \textit{Discriminator model (D)} tries to distinguish the data that is generated by the generator from the real data. The inputs to this model are the real data ($X$) and the generated data ($G(z)$). The output of this model is a binary decision for each data instance, i.e. real/fake.

  \item \textit{Iterative Training:} 
  The GAN is trained using the classification error of the discriminator. The error is used to tune the parameters (weights and biases) of the discriminator, and then the parameters of the generator. Backpropagation~\cite{rumelhart1986learning} is commonly used as the training algorithm. This iterative training consists of two loops:
  
    \begin{itemize}
      \item An \textit{inner loop} where the discriminator's parameters are tuned to maximize the classification accuracy of predicting correct labels for real data and generated data.
      \item An \textit{outer loop} where the generator's parameters are tuned to generate data that has a minimal chance of being distinguished from the real data by the discriminator.
        \end{itemize}
    \end{itemize}
    
The adversarial training of the generator and the discriminator model is a zero-sum game problem: when one model gets better the other one gets worse in equal proportions~\cite{langr2019gans}. For all zero-sum games, there is a point where neither of the players can improve their situation. This point is referred to as the Nash equilibrium. The goal of a GAN is to reach this equilibrium, as then the fake data produced by the generator model is indistinguishable from the real data by the discriminator model. The output of the discriminator is then a random guess on whether the input data is real or fake.

\section{Methodology}\label{sec:Method}

\begin{figure}[tbp]
  \centering
  \includegraphics[width=0.8\linewidth]{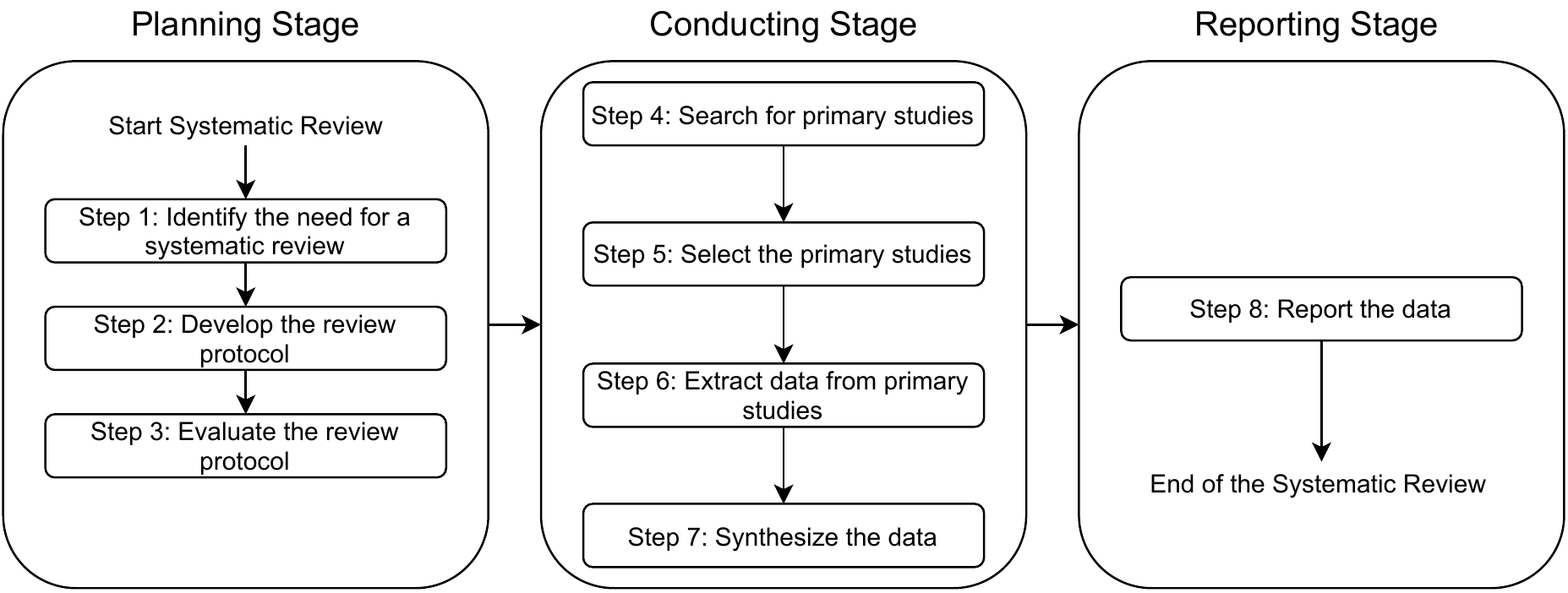}
  \caption{The steps of our systematic literature review, based on Kitchenham's guidelines~\cite{kitchenham}.}
  \label{fig:systematicreview}
  \end{figure}

The planning, conducting, and reporting of this systematic literature review (SLR) were based on the guidelines proposed by Kitchenham~\cite{kitchenham}. The planning stage of the SLR includes three steps: identification of the need for the systematic review, development of the review protocol, and evaluation of the protocol~\cite{kitchenham}. In the conducting stage, based on the review protocol that was developed during planning stage, we search for and select the primary studies, extract data from the primary studies, and synthesize the data. The set of primary studies contains all individual studies that contribute to the SLR~\cite{kitchenham}. In the last stage, we conclude the systematic review by reporting the collected data and findings. Figure~\ref{fig:systematicreview} summarizes the required steps for each stage of the review. In the following, each step is explained in more detail. % However, we avoided quality assessment of the primary studies due to its relatively subjective nature. 

\subsection{The Need for a Systematic Review}\label{sec:objectives}
Recently, GANs have become a hot research topic in many application domains. One of these domains is anomaly detection. The ability of GANs to generate realistic looking data and to perform representation learning makes them attractive for  anomaly detection research. Basically, GANs are trained in an unsupervised manner to learn the distribution of the data. However, they are highly flexible and can be used in semi-supervised fashion as well (e.g.~\cite{springenberg2015unsupervised}). In addition, GANs are implicit density models which do not require any explicit hypothesis on the distribution of the data~\cite{creswell2018generative}. Considering all these advantages, GANs can be leveraged to address some existing problems in anomaly detection, such as the lack of a sufficient amount of data for anomalous behaviour of the system. Therefore, a study summarizing existing research on applications of GANs in anomaly detection would be of a high value to the research community. 

When we started this systematic literature review, we identified only one survey paper~\cite{ADG44} reviewing applications of GANs in anomaly detection. However, this paper only covered 11 papers on anomaly detection with GANs. In addition, it did not follow a systematic approach to conducting the review. This confirmed the need for a systematic literature review on applications of GANs in anomaly detection, which covers a vast number of papers. To reduce researcher bias~\cite{kitchenham}, we  followed a systematic approach to designing, executing and reporting our findings.

\subsection{Developing the Review Protocol}

To reduce the possibility of researcher bias in a systematic manner, a review protocol is required to specify the method for conducting the systematic review~\cite{kitchenham}. This protocol includes definition of the following elements: 1)  research questions, 2) search strategy, 3) study selection criteria (including study quality assessment), 4) data extraction strategy, and 5) synthesis of the extracted data.

% The review protocol is developed and evaluated by regular meetings and discussions in a group of two senior full and assistant professors and two research assistants and improved iteratively during the conducting and reporting stage of this SLR. In the following Subsections ~\ref{sec:ResearchQuestions},~\ref{sec:searchstrategy},~\ref{sec:studyselection}, we present the review protocol in more details. 

% In the first step, we defined a set of research questions based on this SLR's objectives. In the second step, based on the research questions, a search strategy was designed to identify the most relevant primary studies to these research questions. More specifically, in this step, the search terms and the target literature resources are determined. In the third step, a study selection criteria are defined to detect the primary studies that align with the objectives of this SLR and contribute to addressing the defined research questions. The last two steps aim to extract and synthesize from the primary studies. In the data extraction step, a data extraction form is developed to collect the required information in an organized manner from primary studies. Finally, in the synthesizing the data step, we summarized the results of the analysis. 

\subsubsection{Our Research Questions}\label{sec:ResearchQuestions}
In this systematic literature review, we address the following research questions (RQs):
\begin{enumerate}

    \item \textit{RQ1: What is the role of GANs in anomaly detection?} (Section~\ref{sec:RQ1})
    
    \textit{Motivation:} It is important to learn how GANs are used in anomaly detection. One intuitive way is to generate anomalous data to address the problem of the imbalanced dataset. Still, there might be more opportunities. Moreover, we will investigate what are the alternative, non-GAN approaches to handle these identified roles.

   \item \textit{RQ2: What are the application domains of anomaly detection with GANs?} (Section~\ref{sec:RQ2})
   
   \textit{Motivation:} The use of GANs in anomaly detection may be more common in certain domains. Here, we look into which domains and which types of GANs work together well. 

   \item \textit{RQ3: Which GAN architecture is used most often in anomaly detection systems?} (Section~\ref{sec:RQ3})
   
   \textit{Motivation:} There exist many architectures of GANs. Each one attempts to handle a specific type of data or to address an existing problem in the previous architectures. Some architectures may be better suitable for anomaly detection than others. Therefore, we look into which architectures of GANs are commonly used. 
   
   \item \textit{RQ4: Which type of data instance and datasets are most commonly used for anomaly detection with GANs?} (Section~\ref{sec:RQ4})
   
   \textit{Motivation:} Identifying which datasets are used to evaluate anomaly detection with GANs in certain domains can reveal the ``standard benchmarks'' in specific domains and which domains require benchmarks in general. 

   \item \textit{RQ5: Which metrics are used to evaluate the performance of GANs in generating data and anomaly detection?} (Section~\ref{sec:RQ5})
  
    \textit{Motivation:} Evaluating GANs in anomaly detection systems is not a straightforward task as their goal is to create realistic looking data that is different enough from known anomalies, yet still representative of real anomalies. Therefore, one cannot just compare the generated data with the real data. We study which approaches are commonly used to evaluate the quality of the generated data, and support practitioners in deciding which metrics to use for evaluating data in specific anomaly detection problems.  
   
   \item \textit{RQ6: Which anomaly detection techniques are used along with GANs?} (Section~\ref{sec:RQ6})
   
   \textit{Motivation:} GANs are often used together with more traditional anomaly detection techniques, especially when they are used in a supervised manner. In this question, we identify the anomaly detection techniques that are based on or assisted by GANs.
   
\end{enumerate}

  %%%%%  %%%%%  %%%%%  %%%%%  %%%%%  %%%%%  %%%%%  %%%%%  %%%%%  %%%%%  %%%%%  %%%%%  %%%%%  %%%%%  %%%%%  %%%%%  %%%%%  %%%%%  %%%%%  %%%%%  %%%%%  %%%%%  %%%%%  %%%%%  %%%%%  %%%%%  %%%%%  %%%%%  %%%%%  %%%%%  %%%%%  %%%%%  %%%%%  %%%%%  %%%%%  %%%%%  %%%%%  %%%%%  %%%%%  %%%%%  %%%%%  %%%%%  %%%%%  %%%%%  %%%%%  %%%%%  %%%%%  %%%%%  %%%%%  %%%%%  %%%%%  %%%%% 

\subsubsection{Search Strategy}\label{sec:searchstrategy}
To find relevant papers for this systematic review, we searched the IEEE Xplore\footnote{\url{https://ieeexplore.ieee.org}}, ACM Digital Library\footnote{\url{https://dl.acm.org}}, Science Direct\footnote{\url{https://sciencedirect.com}} and Scopus\footnote{\url{https://scopus.com}} digital libraries.
The focus of this study is the application of GANs in anomaly detection. Therefore, we combined the keywords related to anomaly detection with keywords and abbreviations for generative adversarial networks. 
To find closely relevant papers for this study, we searched the title and the abstract of the papers for the following query: \texttt{(``anomaly'' \texttt{OR} ``anomalies'' \texttt{OR} ``anomalous'' \texttt{OR} ``outlier'' \texttt{OR} ``abnormal'') \texttt{AND} (``generative adversarial network'' \texttt{OR} ``generative adversarial networks'' \texttt{OR} ``GAN'' \texttt{OR} ``GANs'')}. The list of primary studies was collected on 3rd June, 2020. 

We conducted a pilot study to ensure that the well-known primary studies were included in the query results. During this study, we searched for the matched papers and their shared references on Google Scholar to ensure that the most cited papers were covered by the query. After several iterations of improvements of the query, we were confident that it returned the important and well-known studies.
%  (\textit{Title} \textbf{OR} \textit{Abstract} :(Anomaly \textbf{OR} Anomalies \textbf{OR} Anomalous \textbf{OR} Outlier \textbf{OR} Abnormal)) \textbf{AND} (\textit{Title} \textbf{OR} \textit{Abstract} :(``Generative Adversarial Networks'' \textbf{OR} ``Generative adversarial Network'' \textbf{OR} GAN \textbf{OR} GANs)). 

% \begin{table}[tbp]
%   \centering
%   \caption{The list of search keywords for primary studies selection.}
%     \begin{tabular}{p{4.75em}cp{13.915em}}
%     \toprule
%     \multicolumn{1}{c}{{Keyword}} & {Operator} & \multicolumn{1}{c}{{Keyword}} \\
%     \midrule
%     Anomaly\newline{}Anomalies\newline{}Anomalous\newline{}Outlier\newline{}Abnormal & AND   & Generative Adversarial Networks\newline{}Generative Adversarial Network\newline{}GAN\newline{}GANs \\
%     \bottomrule
%     \end{tabular}%
%   \label{tab:keywords}%
% \end{table}%

  %%%%%  %%%%%  %%%%%  %%%%%  %%%%%  %%%%%  %%%%%  %%%%%  %%%%%  %%%%%  %%%%%  %%%%%  %%%%%  %%%%%  %%%%%  %%%%%  %%%%%  %%%%%  %%%%%  %%%%%  %%%%%  %%%%%  %%%%%  %%%%%  %%%%%  %%%%%  %%%%%  %%%%%  %%%%%  %%%%%  %%%%%  %%%%%  %%%%%  %%%%%  %%%%%  %%%%%  %%%%%  %%%%%  %%%%%  %%%%%  %%%%%  %%%%%  %%%%%  %%%%%  %%%%%  %%%%%  %%%%%  %%%%%  %%%%%  %%%%%  %%%%%  %%%%%

\subsubsection{Study Selection Criteria}\label{sec:studyselection}
We defined the following criteria for the inclusion of a paper in our study. The last two criteria in the list are included to assess the quality of the study.

\begin{itemize}
\item The paper must be in the specified digital libraries.
\item The primary study should focus on anomaly detection while leveraging GANs.
\item The developed methods should be evaluated on at least one real dataset, not only on simulated data, to ensure practical relevance of the study.
\item The primary study should be available online to ensure accessibility. 
\item The article should be written in the English language. 
\end{itemize}

% \textit{Exclusion Criteria:}
% \begin{itemize}
% \item All the primary studies that do not meet the inclusion criteria.
% \end{itemize}
  \begin{figure}[tbp]
  \centering
  \includegraphics[width=0.5\linewidth]{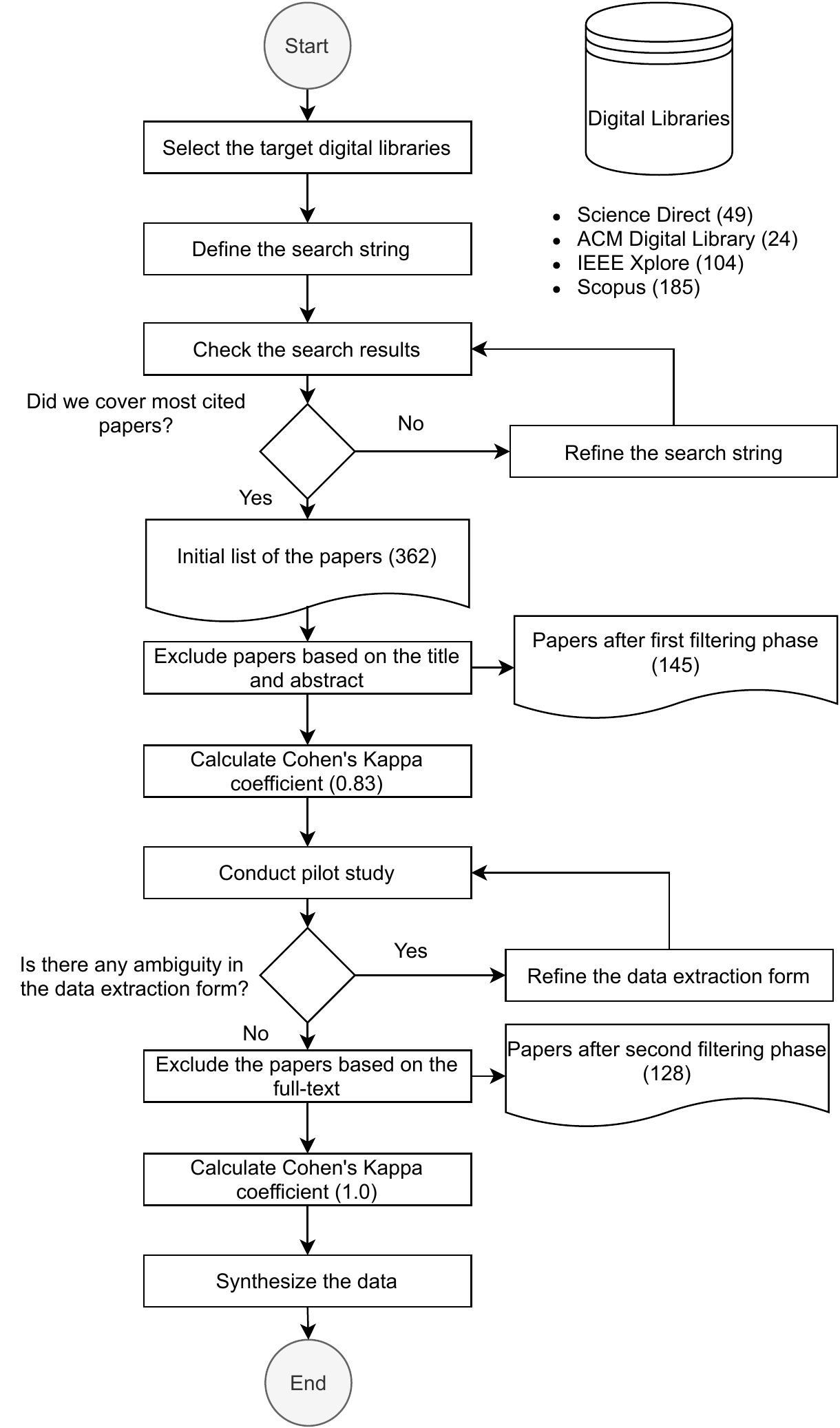}
  \caption{The procedure for searching, selecting, and extracting data from the primary studies for conducting the SLR.}
  \label{fig:searchandselect}
  \end{figure}
  
All types of papers, including journal, conference, workshop, and symposium papers are considered in this review. 
The procedure for searching (described in Section~\ref{sec:searchstrategy}) and selecting the primary studies is shown in Figure~\ref{fig:searchandselect}. The final list of papers used for data extraction and synthesis consists of 128 primary studies.
Not every selected primary study provides answers to all six research questions.
% It is worth pointing out that we avoided quality assessment of the primary studies due to their subjective nature.
  
%   The complete list of selected studies is provided in the appendix section of the paper (Table~\ref{tab:longtable}).

  %%%%%  %%%%%  %%%%%  %%%%%  %%%%%  %%%%%  %%%%%  %%%%%  %%%%%  %%%%%  %%%%%  %%%%%  %%%%%  %%%%%  %%%%%  %%%%%  %%%%%  %%%%%  %%%%%  %%%%%  %%%%%  %%%%%  %%%%%  %%%%%  %%%%%  %%%%%  %%%%%  %%%%%  %%%%%  %%%%%  %%%%%  %%%%%  %%%%%  %%%%%  %%%%%  %%%%%  %%%%%  %%%%%  %%%%%  %%%%%  %%%%%  %%%%%  %%%%%  %%%%%  %%%%%  %%%%%  %%%%%  %%%%%  %%%%%  %%%%%  %%%%%  %%%%%

  %%%%%  %%%%%  %%%%%  %%%%%  %%%%%  %%%%%  %%%%%  %%%%%  %%%%%  %%%%%  %%%%%  %%%%%  %%%%%  %%%%%  %%%%%  %%%%%  %%%%%  %%%%%  %%%%%  %%%%%  %%%%%  %%%%%  %%%%%  %%%%%  %%%%%  %%%%%  %%%%%  %%%%%  %%%%%  %%%%%  %%%%%  %%%%%  %%%%%  %%%%%  %%%%%  %%%%%  %%%%%  %%%%%  %%%%%

\subsubsection{Data Extraction Strategy}\label{sec:dataextractopm}
 To facilitate the data extraction, we devised a data extraction form to collect the required information for each RQ from the primary studies. This form (shown in Table~\ref{tab:dataextractionform}) was refined through several iterations with randomly selected papers on the subject. This refinement was accomplished by comparing the data extraction forms of the two first authors
%  (M.S. and M.Z.) 
 and addressing the potential ambiguities in the data extraction form.

% For example, in RQ5, we had to see the type of anomaly detection techniques used from the data availability perspective. Some of the primary studies do not explicitly specify whether the proposed method is a supervised, semi-supervised or unsupervised anomaly detection technique. In this situation, the research assistants decided the type of anomaly detection technique based on the input data the GAN architecture and after a thorough evaluation of the primary study. 

\begin{table}[tb]
  \centering
  \footnotesize
  \caption{The data extraction form.}
    \begin{tabular}{l}
    \toprule
    % Reviewer Name: \\
    % Review Date: \\
    % Study Identifier: \\
    % Publication Title: \\
    % Names of Authors: \\
    % Publication Source: \\
    % Type of Study: \\
    \textbf{Reviewer Name / 
    Review Date / 
    Study Identifier / 
    Publication Title / 
    Names of Authors / 
    Publication Source / 
    Type of Study} \\
    \textbf{RQ1: What is the role of GANs in anomaly detection?} \\
    \hspace{3mm}Role(s) of the GAN: \\
    \hspace{3mm}Is the GAN used for generating new data or learning the distribution of the data? \\
    \hspace{3mm}What is the type of generated/learned data (normal/abnormal)? \\
    \textbf{RQ2: What are the application domains of anomaly detection with GANs?} \\
    \hspace{3mm}The application domain(s): \\
    \textbf{RQ3: Which GAN architecture is used most often in anomaly detection systems?} \\
    \hspace{3mm}The GAN architecture(s) used in the study: \\
    \textbf{RQ4: Which type of data instance and datasets are most commonly used for anomaly detection with GANs?} \\
    \hspace{3mm}The type of input data to the GAN (main input): (e.g. image, text, etc.) \\
    \hspace{3mm}The preprocessing technique used on the input data: \\
    \hspace{3mm}Datasets that are used for the study: \\
    \hspace{3mm}Usage of the dataset: (e.g., addressing the unbalanced dataset, training on normal/abnormal, etc.) \\
    \textbf{RQ5: Which metrics are used to evaluate the performance of GANs in generating data and anomaly detection?} \\
    \hspace{3mm}The type of performance metrics used for evaluating the performance of GAN: \\
    \textbf{RQ6: Which anomaly detection techniques are used along with GANs?} \\
    \hspace{3mm}The type of anomaly detection techniques used: (e.g., classification-based, clustering-based, etc) \\
    \hspace{3mm}The anomaly detection techniques: (e.g. K- Nearest neighborhood, Neural networks) \\
    \bottomrule
    \end{tabular}%
  \label{tab:dataextractionform}%
\end{table}%

\subsubsection{Data Synthesis}
During the data synthesis step, we aggregate the collected data from the data extraction forms to answer the research questions. Putting all this data together gives invaluable information concerning the current best practices and architectures for anomaly detection with GANs. 

\subsection{Evaluation of the Review Protocol}
The protocol is a critical part of the SLR. It was evaluated by the last two authors
% (C.B. and P.M.) 
and, after several iterations, the final version of the protocol was approved and used throughout the conducting stage of the SLR. 
  %%%%%  %%%%%  %%%%%  %%%%%  %%%%%  %%%%%  %%%%%  %%%%%  %%%%%  %%%%%  %%%%%  %%%%%  %%%%%  %%%%%  %%%%%  %%%%%  %%%%%  %%%%%  %%%%%  %%%%%  %%%%%  %%%%%  %%%%%  %%%%%  %%%%%  %%%%%  %%%%%  %%%%%  %%%%%  %%%%%  %%%%%  %%%%%  %%%%%  %%%%%  %%%%%  %%%%%  %%%%%  %%%%%  %%%%%  %%%%%  %%%%%  %%%%%  %%%%%  %%%%%  %%%%%  %%%%%  %%%%%  %%%%%  %%%%%  %%%%%  %%%%%  %%%%%

% \subsection{Our Contribution}
% This review paper is the first study considering specific RQs to analyse and investigate the application of GANs for anomaly detection in a systematic way to ensure and minimize the possibility of bias toward primary studies.

\subsection{Conducting the SLR}
The conducting stage of the SLR includes the following four steps: searching for primary studies, selecting the primary studies, extracting the primary studies, and synthesizing the data. 
% According to the procedure introduced by Kitchenham et al.~\cite{kitchenham}, we need to assess the quality of the primary studies after selecting the primary studies. However, we skipped this step due to its subjective nature. We will go through the process step by step. 
% \subsection{Search for Primary Studies}

We identified 362 papers that matched our search query (see Section~\ref{sec:searchstrategy}): 49~papers in Science Direct, 24 in ACM Digital Library, 104 in IEEE Xplore, and 185 in Scopus. 
% We documented the search process for each digital library while using their advanced search method, and
We organized the papers for further analysis using Mendeley as a reference manager.

% \subsection{Selecting the Primary Studies}

We filtered out irrelevant and duplicated papers according to the study selection criteria introduced in Section~\ref{sec:studyselection}. Figure~\ref{fig:searchandselect} shows the procedure for selecting the primary studies. We filtered the papers in two steps. In the first step, the first two authors independently read the abstract and the title of the primary studies and decided if they were related to anomaly detection with GANs. The Cohen's kappa coefficient~\cite{mchugh2012interrater} for this binary classification (relevant vs. irrelevant) was 0.83, which shows a satisfactory agreement between the researchers. There were 32 papers from the initial list of papers on which the first two authors disagreed. For those papers, the third author was asked to make the final decision regarding inclusion or exclusion. After the first phase of filtering papers, we ended up with 145 papers. The second filtering step was performed while reading the full text: the first two authors decided to include or exclude the paper in the data extraction step. In this phase, the first two authors made the same decision regarding the excluded papers and excluded 17 papers. Finally, 128 primary studies were included and analysed in this SLR. 
%PMDONE
% \subsection{Extracting Data from Primary Studies}
Based on the data extraction strategy introduced in Section~\ref{sec:dataextractopm}, we examined these 128 primary studies to collect the data that contributes to addressing the RQs of this SLR. The primary studies were randomly divided into 10 batches. For each batch, the first two authors extracted the data from the primary studies and filled in the data extraction forms. After extracting the data from each batch, the data extraction forms were randomly distributed between the first two authors, and then the disagreements were identified and discussed in a meeting. If they failed to reach a consensus, one of the last two authors made the final decision. 
After extracting data from all primary studies and addressing the discrepancies in the data extraction forms, we created a spreadsheet for each data extraction form.
% Since we did not extract any quantitative data from the primary studies, we summarized and reported the data as shown in the following section.%~\ref{sec:Results}.
We summarize and report the extracted data in the following section.

% \subsection{Assess the Quality of the Primary Studies}

% \subsection{Synthesizing the Data}
% In this step, we first created a spreadsheet of extracted data for each research question.\textbf{TALK TO CP}

% \subsection{Reporting the Data}
% In this step we decided how to present the results of the SLR.\textbf{TALK TO CP}
\section{Results}\label{sec:Results}

%This section presents and discusses the findings of this systematic literature review.
%First, we present the overview of the selected primary studies. Second, we report and discuss our findings according to the research questions, one by one in the separate subsections. Section~\ref{sec:overviewselected} presents an overview of the selected primary studies. In Section~\ref{sec:RQ1} we discuss our findings regarding the different types of GANs used in the primary studies. Section~\ref{sec:RQ2} is concerned with discussing the application domains of GANs for anomaly detection. Section~\ref{sec:RQ3} discusses the datasets used for anomaly detection in the primary studies. In Section~\ref{sec:RQ4} the roles of GAN in anomaly detection techniques are discussed. Section~\ref{sec:RQ5} presents the performance evaluation metrics for GANs used in the primary studies, and Section~\ref{sec:RQ6} discusses all the GAN based anomaly detection techniques along with other non-GAN anomaly detection techniques which can be seen as the alternatives. 

\begin{figure*}[tbp]
\begin{subfigure}{.5\textwidth}
  \centering
  \includegraphics[width=0.9\linewidth]{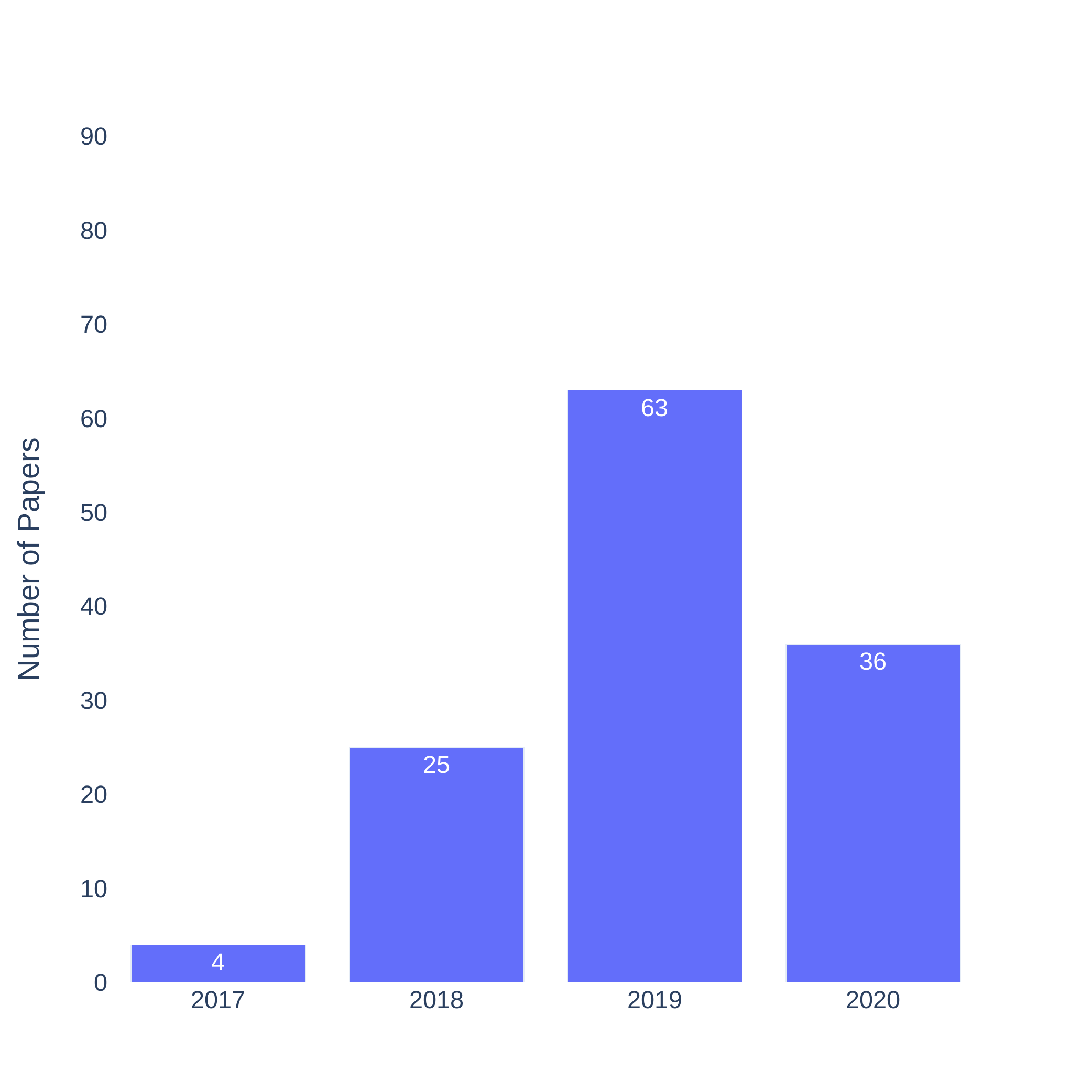}
  \caption{Publication years.}
  \label{fig:PublicationYears}
\end{subfigure}%
\begin{subfigure}{.5\textwidth}
  \centering
  \includegraphics[width=0.9\linewidth]{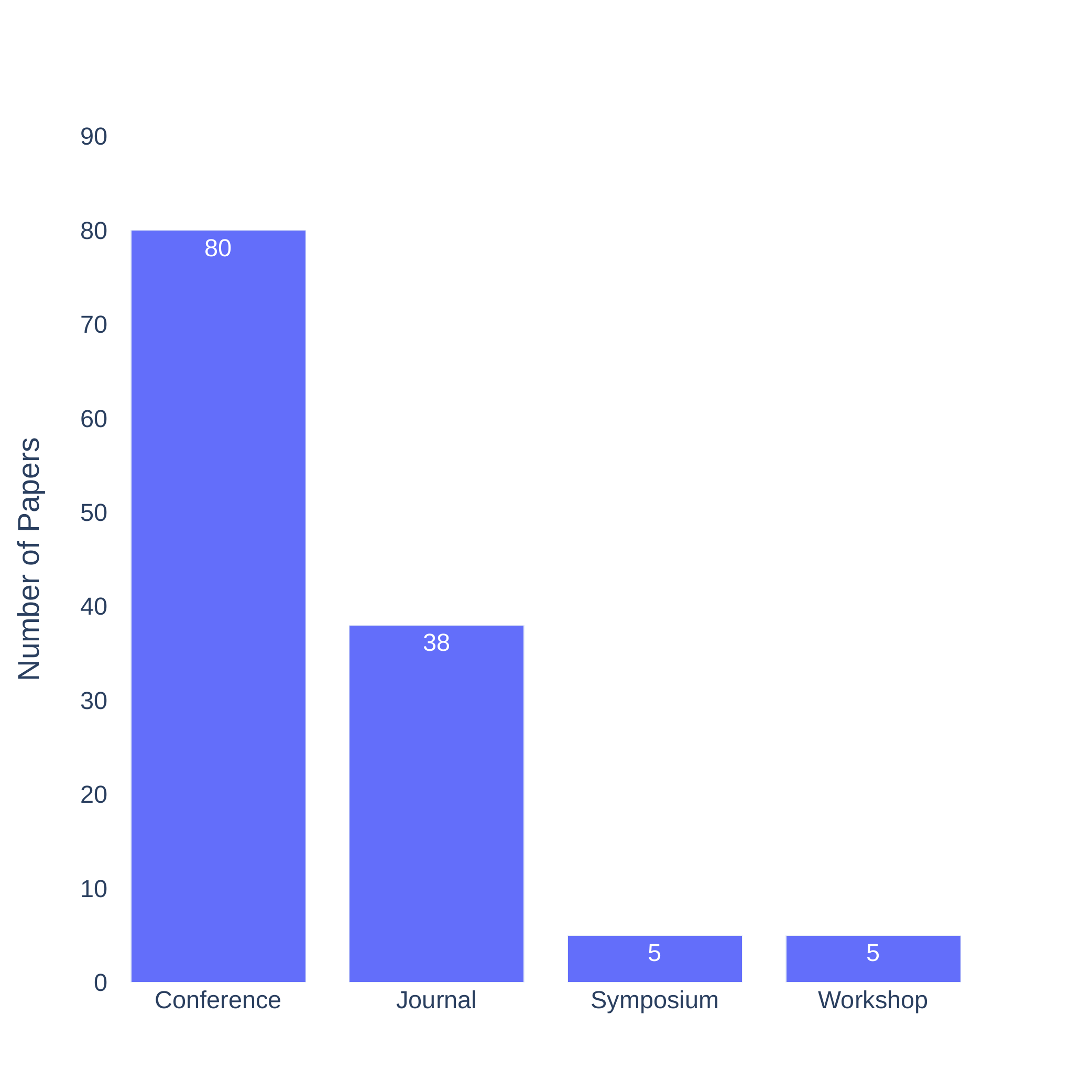}
  \caption{Publication types.}
  \label{fig:PublicationTypes}
\end{subfigure}
\caption{Distribution of the primary studies according to the publication types and years.}
\label{fig:TypeandYear}
\end{figure*}

%\subsection{Overview of the Selected Studies}
\label{sec:overviewselected}
This systematic literature review covers 128 primary studies that describe applications of GANs in anomaly detection. As shown in Figure~\ref{fig:PublicationYears} these primary studies were published between 2017 and early 2020. The number of studies per year is increasing, suggesting that interest in this research area is growing rapidly. Figure~\ref{fig:PublicationTypes} shows that the majority of the reviewed papers (63\%) appeared in conference proceedings, 29\% of the papers were published in journals, and 4\%  in workshops and 4\% in symposia.

% Figure~\ref{fig:publishertypes} shows that, 59.2\% of the primary studies are published by \text{IEEE} and 14.7\% of them are published by \textit{Springer}.
The main publication venues include \textit{IEEE Access}, \textit{IEEE International Symposium on Biomedical Imaging (ISBI)}, \textit{IEEE International Conference on Acoustics, Speech and Signal Processing (ICASSP)}, and \textit{IEEE Conference on Computer Vision and Pattern Recognition}. We found 108 different venues, which shows that the primary studies are not concentrated in a single specific journal or conference.
%
% Regarding the types of the selected primary studies, they all describe experimental research, with only one study~\cite{ADG44} presenting a survey.

%   \begin{figure}[tbp]
%   \centering
%   \includegraphics[width=0.8\linewidth]{Publisher_Types.pdf}
%   \caption{Distribution of the Primary Studies According to the Publishers.}
%   \label{fig:publishertypes}
%   \end{figure}

% Lack of case studies shows that the application of GANs for anomaly detection in the industry is still immature and yet to develop. 
\subsection{RQ1: What is the role of GANs in anomaly detection?}\label{sec:RQ1}
%In this section, we discuss have GANs been used in anomaly detection. After investigating the primary studies, we 
We identified four types of GAN applications in anomaly detection: (1) generating abnormal data instances, (2) generating normal and abnormal instances, (3) learning the normal behaviour of a system, (4) learning both normal and abnormal behaviour of a system. Applications 1 and 2 can be classified as data augmentation with GANs and 3 and 4  as representation learning with GANs. 

Generative models, such as GANs, are mainly designed for data augmentation, i.e., to generate new data and use it to augment the existing data. They can also be used for representation learning, i.e., to learn representations of the data to support information extraction for use when building classifiers or other predictors~\cite{bengio2014representation}. In this case, the generator and the discriminator of a GAN can be used to learn the distribution of a specific class of data, i.e. normal or abnormal data. In turn, the learned distribution can be used to identify nonconforming or irregular data. Table~\ref{tab:TableTask} shows the two main roles of GANs in anomaly detection, along with the types of data used in each role. 
Most of the primary studies opted to use GANs to learn the representation of the data rather than augmenting the datasets. Moreover, this representation learning is mainly performed on normal data. The rationale behind this preference is that, due to the data imbalance, it is usually easier to learn a model of normality rather than abnormality. In addition, by learning only the normal data distribution, the need for data from the abnormal condition of the target system is eliminated. 
% Surprisingly, the table reveals that GANs are more commonly used for representation learning than for data augmentation. %In the following subsections, we will discuss the research efforts carried out to use GANs for data augmentation and representation learning, which the summary of this classification is shown in Table~\ref{tab:TableTask}.

% \begin{figure}[tbp]
% \centering
% \includegraphics[width=0.75\linewidth]{GAN_Tasks.pdf}
% \caption{Distribution of different data types used for data augmentation and representation learning with GANs.}
% \label{fig:GANTasks}
% \end{figure}

\begin{table}[tb]
  \centering
  \caption{List of studies using normal, abnormal, and normal and abnormal data together for different tasks of GANs.}
    \begin{tabular}{llp{0.5\columnwidth}}
    \toprule
    \multicolumn{1}{l}{{\textbf{Task}}} & \textbf{Data type} & \textbf{List of references}\\
    \midrule
    
    \multirow[t]{2}{*}{Representation Learning} & Normal & \cite{ADG03,ADG04,ADG06,ADG07,ADG08,ADG09,ADG11,ADG12,ADG13,ADG14,ADG18,ADG21,ADG22,ADG23,ADG24,ADG25,ADG26,ADG29,ADG30,ADG31,ADG32,ADG33,ADG34,ADG35,ADG36,ADG37,ADG38,ADG39,ADG40,ADG41,ADG42,ADG43,ADG45,ADG46,ADG47,ADG48,ADG49,ADG50,ADG51,ADG52,ADG54,ADG55,ADG56,ADG57,ADG58,ADG60,ADG61,ADG62,ADG63,ADG64,ADG65,ADG66,ADG68,ADG71,ADG72,ADG73,ADG75,ADG77,ADG78,ADG79,ADG80,ADG81,ADG82,ADG85,ADG87,ADG90,ADG93,ADG94,ADG95,ADG97,ADG98,ADG99,ADG100,ADG101,ADG102,ADG103,ADG105,ADG108,ADG109,ADG110,ADG113,ADG114,ADG115,ADG116,ADG117,ADG119,ADG121,ADG127,ADG128,ADG129} \\
              & Normal and Abnormal & \cite{ADG76,ADG118,ADG120,ADG125} \\
    \midrule
    \multirow[t]{2}{*}{Data Augmentation} & Abnormal & \cite{ADG01,ADG02,ADG05,ADG10,ADG15,ADG16,ADG17,ADG19,ADG20,ADG27,ADG53,ADG59,ADG67,ADG69,ADG70,ADG74,ADG83,ADG84,ADG88,ADG92,ADG96,ADG106,ADG122,ADG123,ADG124,ADG126} \\
          & Normal and Abnormal & \cite{ADG28,ADG86,ADG89,ADG91,ADG104,ADG107,ADG111,ADG112} \\
    \bottomrule
    \end{tabular}%
  \label{tab:TableTask}%
\end{table}%

\subsubsection{Representation Learning With Generative Adversarial Networks}

\begin{table}[tbp]
  \centering
  \caption{Representation learning with GANs}
     \begin{tabular}{lm{39em}}
    \toprule
    \textbf{Type of GAN} & \textbf{List of references} \\
    \midrule
    
    DCGAN & \cite{ADG11,ADG14,ADG25,ADG26,ADG29,ADG31,ADG36,ADG40,ADG46,ADG50,ADG52,ADG58,ADG71,ADG81,ADG90,ADG98,ADG100,ADG105,ADG110,ADG114,ADG125,ADG127,ADG128,ADG129,ADG06,ADG24,ADG65,ADG82,ADG95,ADG116,ADG120,ADG113}  \\
    Standard GAN & \cite{ADG03,ADG09,ADG13,ADG32,ADG34,ADG39,ADG41,ADG42,ADG48,ADG49,ADG51,ADG56,ADG57,ADG60,ADG72,ADG79,ADG85,ADG94,ADG99,ADG108,ADG109,ADG117,ADG118}  \\
    cGAN  & \cite{ADG12,ADG18,ADG33,ADG35,ADG54,ADG55,ADG63,ADG64,ADG75,ADG77,ADG78,ADG80,ADG97,ADG102}  \\
    BiGAN & \cite{ADG22,ADG30,ADG43,ADG47,ADG61,ADG66,ADG99,ADG103}  \\

    % AnoGAN & \cite{ADG06,ADG24,ADG65,ADG82,ADG95,ADG116,ADG120} \\
    WGAN  & \cite{ADG04,ADG21,ADG66,ADG87,ADG99,ADG110}  \\
    WGAN-GP & \cite{ADG38,ADG62,ADG76,ADG121}  \\
    VAE-GAN & \cite{ADG07,ADG08,ADG115,ADG119}  \\    
    O-GAN & \cite{ADG23,ADG73} \\    
    Cycle-GAN & \cite{ADG45}  \\
    EBGAN & \cite{ADG54} \\
    GAN-QP & \cite{ADG73} \\
    OCGAN & \cite{ADG101} \\
    PatchGAN & \cite{ADG37} \\
    RaSGAN & \cite{ADG93} \\
    TextGAN & \cite{ADG68} \\
    \bottomrule
    \end{tabular}%
  \label{tab:RepresentationLearning}%
\end{table}%

The main goal of GANs is to learn a generative model that produces realistic-looking data by sampling from the learned distribution.
This generative power of GANs was highlighted by Goodfellow et al.~\cite{GANGoodfellow} and Radford et al.~\cite{DCGANradford}. Representation learning with GANs for anomaly detection exploits the ability of GANs in learning the distribution of a specific class of data. Several anomaly detection techniques are proposed that use this representation learning ability of GANs (shown in Table~\ref{tab:RepresentationLearning}). We will explain the concept of anomaly detection using representation learning through an examples of a well-known GAN-based anomaly detection techniques (AnoGAN). All other anomaly detection techniques that rely on representation learning through a GAN are variations to some extent of the AnoGAN technique.

Schlegl et al.~\cite{ADG128} introduced the first GAN-based anomaly detection technique, called AnoGAN, taking advantage of the representation learning ability of GANs . Schlegl et al. put forth AnoGAN, which employs the DCGAN architecture, to learn the distribution of normal anatomical variability. The idea comes from the concept of a smooth transition in the latent space of the data, i.e., that sampling from two close points in the latent space should lead to similar data in the data space~\cite{yeh2017semantic}.
Schlegl et al. hypothesize that the latent vector of the GANs represents the distribution of the trained data. Therefore, one can learn the representation of the normal data by training GANs only on normal data. From an anomaly detection view, learning the representation of the normal data is useful as one can decide for new (potentially anomalous) data points how likely they are part of that normal data. During the training of a GAN, the generator learns the mapping from latent space to data space $G(z)=z \rightarrow x$ (i.e., the representation of the data). However, the inverse mapping, which is necessary to decide whether a data point is anomalous, is not straightforward to obtain~\cite{ADG128}. 
To address this problem, Schlegl et al. proposed an additional step after training the GAN on normal data. For an image $x$, they proposed to find a point $z$ in the latent space that corresponds to an image $G(z)$, which is the most similar to the image $x$ on the learned manifold $\chi$. Schlegel et al. proposed an iterative process to find the most similar image $G(z_{\Gamma})$ to $x$ using residual and discrimination loss. The similarity of images $x$ and $G(z)$ depends on how closely $x$ follows the distribution of the data learned by the generator $(p_{g})$. After identifying the most similar image, AnoGAN computes an anomaly score that is related to the similarity of $x$ and $G(z)$. Finally, based on a threshold for the anomaly score, AnoGAN decides whether $x$ is an anomaly.

\subsubsection{Data Augmentation with Generative Adversarial Networks}
Machine learning techniques, especially deep learning methods~\cite{marcus2018deep}, require a massive amount of data to perform well in their designated task~\cite{obermeyer2016predicting}. Data augmentation, also known as oversampling, is carried out to compensate for an insufficient amount of data in the dataset to prevent model overfitting. It can also be used to address the problem of data imbalance, which occurs when the sizes of the classes in a dataset differ considerably. 
For instance, in a binary classification task, the class with fewer samples is called the minority class, and the other class is called the majority class. The corresponding training process would be biased towards the majority class, hence a classifier trained using this dataset would have a better accuracy for this class~\cite{kaur2019systematic}. To address the imbalanced dataset problem, one can either randomly remove samples from the majority class to balance the class size (undersampling), or augment the minority class by adding artificially generated instances (oversampling) using suitable techniques. 

The problem of the imbalanced dataset is more critical in anomaly detection since it is hard and expensive to collect data on anomalous behaviour of the system under study. Often, there are very few or no examples of anomalous data available. In this situation, GANs can help by generating more samples for the anomalous class. 

% Generative adversarial networks is an oversampling method that tries to learn the statistical distribution of the training data. This enables the generator of the GANs to generate samples from the learned distribution of the data and consequently augment the generated data with the available data. Goodfellow et al.~\cite{GANGoodfellow} showed the generative adversarial networks' superiority in generating synthetic data learned from the training dataset. 

Table~\ref{tab:TableTask} summarizes the main roles of GANs in anomaly detection and the type of data used for that purpose. Data augmentation with GANs is mostly used to generate data that represents anomalous behaviour of the system. There was no primary study augmenting only the normal condition of the system under study for anomaly detection. This is due to the fact that there is usually an abundance of data for the normal condition. However, some studies augmented both normal and abnormal data, e.g., using CycleGANs, by learning the transformation from abnormal to normal and from normal to abnormal to generate new data. After augmenting the dataset, it is ready to be used for anomaly detection, usually performed by a classifier (as discussed in Section~\ref{sec:RQ6}). Most primary studies that use GANs for data augmentation report a slight improvement in classification accuracy compared either to traditional techniques or without data augmentation.
For instance, Madani et al.~\cite{ADG126} report that, using data augmentation with GANs, the test accuracy for cardiovascular abnormality detection improved from 81.93\% to 84.19\%. In comparison, using traditional augmentation methods, they achieved only 83.12\% test accuracy. This improvement is significant when dealing with large amounts of data, especially in medical applications. However, in some studies, it has been reported that GAN did not meet their expectations in improving the classification accuracy after augmenting the data, e.g.~\cite{ADG107}. The list of different GANs used for data augmentation is shown in Table~\ref{tab:DATable}.

\begin{table}[tbp]
  \centering
  \caption{Different types of GANs used for data augmentation.}
    \begin{tabular}{llll}
    \toprule
    \textbf{Type of GAN} & \multicolumn{1}{p{12em}}{\textbf{List of references}} & \textbf{Type of GAN} & \multicolumn{1}{p{15em}}{\textbf{List of references}} \\
    \midrule
    AAE   & \cite{ADG122} & D2GAN & \cite{ADG70} \\
    AC-GAN & \cite{ADG17,ADG20,ADG104} & DCGAN & \cite{ADG28,ADG67,ADG74,ADG84,ADG92,ADG107,ADG111,ADG112,ADG126} \\
    BGAN  & \cite{ADG17} & PG-GAN & \cite{ADG17,ADG89} \\
    BiGAN & \cite{ADG16} & SeqGAN & \cite{ADG59} \\
    cGAN  & \cite{ADG01,ADG02,ADG69,ADG91,ADG106,ADG123,ADG124} & Standard GAN & \cite{ADG05,ADG10,ADG15,ADG27,ADG53,ADG88,ADG96} \\
    Cycle-GAN & \cite{ADG19,ADG70,ADG86} & WGAN  & \cite{ADG83} \\
    \bottomrule
    \end{tabular}%
  \label{tab:DATable}%
\end{table}%

In the examined primary studies, we identified several traditional techniques for addressing the problem of imbalanced and insufficient amounts of data. The effects of adopting these techniques are compared to the GANs in terms of improving the classification accuracy.
For example, random undersampling was evaluated in two primary studies~\cite{ADG01,ADG83}, where samples were randomly removed from the majority class. Using this approach, some important and critical data may be lost that could otherwise be beneficial for learning a robust decision boundary~\cite{Imbalanced_Learning}. Random oversampling was investigated in four primary studies~\cite{ADG01,ADG02,ADG50,ADG83}. In this case, some samples from the minority class are copied to increase its size. However, this approach is likely to cause over-fitting~\cite{ADG83}. All these studies confirmed the superiority of GANs in data augmentation compared to random over/undersampling.

Chawla et al.~\cite{chawla2002smote} proposed synthetic minority oversampling (SMOTE) to improve the random oversampling by synthesizing new samples from the neighbourhood of the minority class samples. This improvement is accomplished by interpolating between several minority class instances. SMOTE and its variants (e.g., borderline-SMOTE~\cite{han2005borderline}) were compared with GANs in several studies~\cite{ADG01,ADG02,ADG19,ADG50,ADG83,ADG122}.
%Another variant of SMOTE is borderline-SMOTE put forth by Han et al.~\cite{han2005borderline} and compared to GANs in terms of performnce by two primary studies~\cite{ADG02,ADG50}. Moreover, some other flavors of SMOTE, such as SMOTE+Tomek links and SMOTE+EEN~\cite{batista2004study} have been investigated by Jiang et al.~\cite{ADG50}. 
The ADAptive SYNthetic (ADASYN) sampling approach for imbalanced datasets~\cite{he2008adasyn} was compared with GANs in two studies~\cite{ADG50,ADG83}. ADASYN uses a weighted distribution for different minority class instances based on their difficulty level, i.e., the more difficult to learn instances are synthesized more frequently. %Authors of~\cite{ADG50,ADG83} compared this type of data augmentation with the augmentation capabilities of GANs. 
In addition, several other traditional techniques for data augmentation, such as adding Gaussian noise to the dataset~\cite{ADG20}, Gaussian smoothing, unsharp masking, minimum filtering~\cite{ADG89}, and affine transforms~\cite{ADG107} were compared to GANs. Most of these studies show that data augmentation with GANs results in training datasets that improve the anomaly detection.

\subsection{RQ2: What are the application domains of anomaly detection with GANs?}\label{sec:RQ2}
 Table~\ref{tab:ApplicationDomains} shows the different domains where GANs were applied in the primary studies. The table reveals that a vast number of primary studies (24 papers) perform anomaly detection in medical applications, closely followed by surveillance and intrusion detection with 19 and 17 papers, respectively.
 %were the other domains in which GANs were used for anomaly detection. We will briefly introduce the identified application domains in more detail. 

% \begin{figure}[htbp!]
% \centering
% \includegraphics[width=1\linewidth]{GAN_Applications.pdf}
% \caption{Distribution of applications of generative adversarial networks.}
% \label{fig:GANapplications}
% \end{figure}

\begin{table}[htbp]
  \centering
  \small
  \caption{The application domains of GANs for anomaly detection}
     \begin{tabular}{lm{40em}}
    \toprule
    \multicolumn{1}{l}{{\textbf{Application domain}}} & \textbf{List of references} \\
    \midrule
    Medical & ~\cite{ADG04,ADG06,ADG18,ADG23,ADG24,ADG38,ADG57,ADG69,ADG82,ADG86,ADG87,ADG89,ADG97,ADG107,ADG112,ADG114,ADG115,ADG119,ADG122,ADG123,ADG124,ADG126,ADG128,ADG129} \\
    Surveillance & ~\cite{ADG08,ADG09,ADG11,ADG13,ADG25,ADG33,ADG35,ADG37,ADG39,ADG51,ADG63,ADG64,ADG67,ADG78,ADG79,ADG90,ADG102,ADG103,ADG125} \\
    Intrusion Detection  & ~\cite{ADG01,ADG05,ADG07, ADG15, ADG19,ADG27,ADG42,ADG59,ADG76,ADG80,ADG88,ADG92,ADG93,ADG94,ADG99,ADG106,ADG108} \\
    \multicolumn{1}{l}{Various} & ~\cite{ADG02,ADG34,ADG47,ADG48,ADG49,ADG53,ADG61,ADG66,ADG73,ADG85,ADG113,ADG118} \\
    System Health & ~\cite{ADG03,ADG17,ADG22,ADG50,ADG55,ADG56,ADG74,ADG91,ADG105,ADG117} \\
    Image Recognition & ~\cite{ADG14,ADG28,ADG43,ADG48, ADG52, ADG62,ADG65,ADG98, ADG100, ADG101, ADG110} \\
    Manufacturing & ~\cite{ADG29,ADG36,ADG40,ADG70,ADG71,ADG81,ADG116,ADG120} \\
    Autonomous Systems & \multicolumn{1}{l}{~\cite{ADG12,ADG20,ADG25,ADG54,ADG58,ADG75,ADG77,ADG95}} \\
    Power/Energy & \multicolumn{1}{l}{~\cite{ADG26,ADG31,ADG45,ADG46,ADG84,ADG121}} \\
    Fraud Detection & ~\cite{ADG10, ADG41, ADG83, ADG96, ADG127} \\
    Hyperspectral Images & ~\cite{ADG21, ADG32, ADG72, ADG104} \\
    Trajectory Detection & ~\cite{ADG16,ADG60} \\
    Software Systems & ~\cite{ADG30, ADG109} \\
    Text  & ~\cite{ADG68} \\
    Climate Changes & ~\cite{ADG111} \\
    \bottomrule
    \end{tabular}%
  \label{tab:ApplicationDomains}%
\end{table}%


\textit{Medical Anomaly Detection.}
Anomaly detection in medicine deals with analyzing patients' health conditions using medical records and images~\cite{anomalysurvey}. Specific applications include retinal optical coherence tomography (OCT) anomaly detection~\cite{ADG04,ADG18,ADG23,ADG128}, seizure detection~\cite{ADG06}, cardiovascular disease detection~\cite{ADG24}, lung nodule detection~\cite{ADG38}, abnormal chest X-ray identification~\cite{ADG57,ADG89,ADG114,ADG126}, polyp detection~\cite{ADG69,ADG87}, metastatic bone tumor detection~\cite{ADG82}, lesion detection~\cite{ADG86,ADG119}, laparoscopy anomaly detection~\cite{ADG97}, breast cancer detection~\cite{ADG112,ADG122}, MRI quality control~\cite{ADG115}, diabetic retinopathy detection~\cite{ADG123}, brain tumor detection~\cite{ADG124} and hemorrhage detection~\cite{ADG129}.
One of the challenges in this domain is the difficulty of obtaining expert labels for medical data, such as clinical images, since annotation is an exhaustive and time-consuming task.

\textit{Surveillance Anomaly Detection.}
To improve public safety, surveillance cameras are widely used in public places such as streets, stores, and banks. The goal of video surveillance is to identify suspicious activity, unusual traffic patterns, or accidents by automatically analyzing the behaviour of the surveillance target. In video surveillance anomaly detection, this can be accomplished by identifying the out-of-ordinary behaviours that differ from dominant (normal) behaviours in the scene~\cite{anomalysurvey}. Automated video surveillance can reduce the dependence on human workers and reduce the risk of late detection of anomalous behaviour. Most primary studies in this application domain leverage GANs for video anomaly detection to find irregularities in the crowds. However, traffic anomaly detection~\cite{ADG13} and threat object recognition with X-ray imaging~\cite{ADG103} have also been studied. 

\textit{Intrusion Detection.}
Intrusion detection systems are defined as software and/or hardware components that monitor and analyze events in computer systems to identify signs of intrusion~\cite{lazarevic2005intrusion}. Any malicious intrusion or attack on network vulnerabilities, computers or information systems may result in a serious predicament and violate the confidentiality, integrity and availability of the systems~\cite{liao2013intrusion}. The examined primary studies are mainly focused on network intrusion detection~\cite{ADG01,ADG19,ADG27,ADG59,ADG76,ADG88,ADG92,ADG94,ADG99,ADG106}. However, other applications of GANs in intrusion detection are smartphone lock pattern intrusion detection~\cite{ADG05}, presentation attack detection~\cite{ADG93}, phishing detection~\cite{ADG15}, cognitive radio intrusion detection~\cite{ADG80}, cyber-physical system intrusion detection~\cite{ADG108}, and IoT security~\cite{ADG07,ADG42}.

\textit{General Approaches.}
Some primary studies do not focus on a single application domain. Instead, they evaluate the proposed approaches in different application domains (shown as \emph{Various} in Table~\ref{tab:ApplicationDomains}).
For example, three primary studies~\cite{ADG34,ADG47,ADG48} investigate their proposed GAN-based anomaly detection for intrusion detection and image recognition. 
Two evaluated studies~\cite{ADG85,ADG118} apply GAN-based anomaly detection in image recognition and video surveillance applications. Other primary studies evaluate their anomaly detection approach in intrusion detection, medical and image recognition domains~\cite{ADG61,ADG73}. Khoshnevisian et al.~\cite{ADG49} investigate the application of their proposed GAN-based anomaly detection in medicine and on trajectory anomaly detection. Hyuk et al.~\cite{ADG02} evaluate their proposed technique for image recognition in addition to medical and trajectory anomaly detection. Wang et al.~\cite{ADG66} study the application of GANs in fraud and intrusion detection,
and Liu et al.~\cite{ADG53} evaluate their approach in medicine, image recognition, aviation, human activity, spam identification, and waveform anomaly detection.

\textit{System Health Anomaly Detection.}
System health monitoring is a way to identify anomalous behaviour in large (often industrial) systems.%, such as  railway systems, communication systems, or storage systems. 
In industrial processes, the anomalous behaviour can represent, for example, wear or damage to the industrial equipment after continuous use. It is critical that such degradations in a system's performance are detected before they escalate and cause loss of revenue or endanger human life. Examples of industrial applications of system health anomaly detection with GANs include industrial process anomaly detection~\cite{ADG03,ADG22}, electrical insulator anomaly detection~\cite{ADG17}, rolling bearing anomaly detection~\cite{ADG50}, steam turbine anomaly detection~\cite{ADG56}, magnetic flux leakage detection~\cite{ADG91}, fused magnesium furnace anomaly detection~\cite{ADG74}, railway turnout anomaly detection~\cite{ADG105} and communication system anomaly detection~\cite{ADG117}.

\textit{Image Recognition.}
Image anomaly detection refers to finding images with abnormal patterns that do not comply with other images in the same set. Most primary studies in this application domain use public image datasets, such as MNIST or CIFAR-10, to prove the concept of their proposed anomaly detection techniques. However, Moussa et al.~\cite{ADG28} evaluate the application of GANs for object recognition in images, such as finding an airplane in the picture. Bergmann et al.~\cite{ADG65} propose a dataset of high-resolution color images of different object and texture categories suitable for anomaly detection.
% defects in natural images. 
They evaluate several anomaly detection techniques, including GANs, to process their dataset. The proposed dataset aims to provide more challenging images than the commonly used datasets mentioned above. 

\textit{Manufacturing Anomaly Detection.}
This anomaly detection application refers to the quality inspection of manufactured products to identify defective products. These defects reveal themselves as irregularities on metal or wood surfaces, electronic parts, and so on. For example, an application of visual surface defect detection is studied in four primary studies~\cite{ADG29,ADG40,ADG70,ADG71} and industrial quality inspection is investigated in three studies~\cite{ADG36,ADG81,ADG116}.

\textit{Anomaly Detection in Autonomous Systems.}
Autonomy is defined as self-governance or freedom from external influences~\cite{nguyen2012evolutionary}.
An autonomous system is referred to as a system that can perceive the environment, make decisions based on the sensed information, and then react to internal/external changes using actuators. However, a fault may occur in each of these steps. 
For example, in an autonomous robot, faults can occur in sensors, software, or after physical damage to the actuator. This domain includes driving anomaly detection~\cite{ADG12,ADG20,ADG95} to assist the driver or to identify abnormalities in the driver's behaviour. Autonomous surveillance with moving agents is addressed in three primary studies~\cite{ADG25,ADG75}, where an autonomous moving agent, such as a patrol robot, scans the environment to find abnormal activity. Two primary studies focused on controller anomaly detection~\cite{ADG54,ADG77}, to identify abnormal decision making by a controller in a closed-loop control system. Sun et al.~\cite{ADG58} study autonomous vehicle anomaly detection.

\textit{Power/Energy Anomaly Detection.}
This application domain is concerned with identifying abnormalities in the power/energy consumption and power/energy infrastructure. Examples include catenary support component anomaly detection~\cite{ADG26,ADG31,ADG46}, power plant anomaly detection~\cite{ADG45,ADG84} and power consumption anomaly detection~\cite{ADG121}.

\textit{Fraud Detection.}
Fraud is defined as exploiting one's occupation for personal enrichment by willful misuse or application of their employer's resources or assets without authorization~\cite{kou2004survey}. Fraud detection refers to uncovering these illegal activities. Examples of applications in this domain include click advertisement fraud~\cite{ADG10}, stock market manipulation~\cite{ADG41}, credit card fraud~\cite{ADG83}, health care insurance providers fraud~\cite{ADG96}, and satellite image forgery~\cite{ADG127}.

\textit{Other Domains of Anomaly Detection with GANs.}
There are several additional application domains for anomaly detection using GANs that are less common: trajectory anomaly detection~\cite{ADG16,ADG60}, human mobility anomaly detection~\cite{ADG16}, climate change~\cite{ADG111}, text anomaly detection~\cite{ADG68}, and software systems anomaly detection~\cite{ADG109,ADG30}.

\subsection{RQ3: Which GAN architecture is used most often in anomaly detection systems?}\label{sec:RQ3}
Many types of GANs have been proposed to tackle the deficiencies of the first type of GAN proposed by Goodfellow et al.~\cite{GANGoodfellow} or to handle specific tasks. 
In most cases, they modify the GAN architecture or the cost function of the generator and discriminator.
According to the GAN Zoo GitHub repository\footnote{\url{https://github.com/hindupuravinash/the-gan-zoo}}, more than 500 types of GAN were identified from 2014 to 2018.

We identified 21 different types of GANs used for anomaly detection purposes (see Table~\ref{tab:GANTypes} for a list of primary studies using each of these architectures). DCGANs, standard GANs, and cGANs are the most commonly used GAN architectures. These were among the first proposed GAN architectures, and there are many new ones which are not (yet) used for anomaly detection purposes. The correspondence between the identified GAN architectures and their application domains is shown in Table~\ref{tab:archvsapp}. DCGANs, standard GANs, and BiGANs have been used in various application domains, indicating their flexibility. A variety of GAN architectures have also been used for applications in medicine, intrusion detection, and system health. However, some of the application domains are not well researched regarding GAN architectures, such as text anomaly detection and fraud detection. 

Since the anomaly detection techniques examined in this review are either based on or assisted by GANs, any deficiency in the networks used for anomaly detection directly impacts the performance of the corresponding anomaly detection techniques. Therefore, the improvements in anomaly detection techniques using GANs are strongly correlated with the advances in the GAN architecture and training strategies. 
There are many studies in the literature describing the challenges of existing GANs and available solutions~\cite{pan2019recent,gui2020review,jabbar2020survey,creswell2018generative}. The most crucial problem with GANs is the problem of mode collapse. When this happens, the generator of a GAN always generates samples from a highly concentrated distribution (partial collapse)~\cite{creswell2018generative}, or simply a single sample (complete collapse)~\cite{arjovsky2017towards,arora2017generalization}. Therefore, the generated data lacks the expected diversity. There have been several treatments proposed to lessen the effect of mode collapse during GAN training, such as WGAN~\cite{WGANArjovsky}, and Unrolled GAN~\cite{salimans2016improved}. Another challenge of training of GANs is the instability of the training process and its failure to converge to a Nash equilibrium. Methods proposed to address this problem include Two Time-Scale Update Rule (TTUR)~\cite{heusel2017gans}, WGAN~\cite{WGANArjovsky} and feature matching~\cite{salimans2016improved}.

\begin{figure*}[bt]
\begin{subfigure}[b]{.48\textwidth}
  \centering
  \includegraphics[width=1\linewidth]{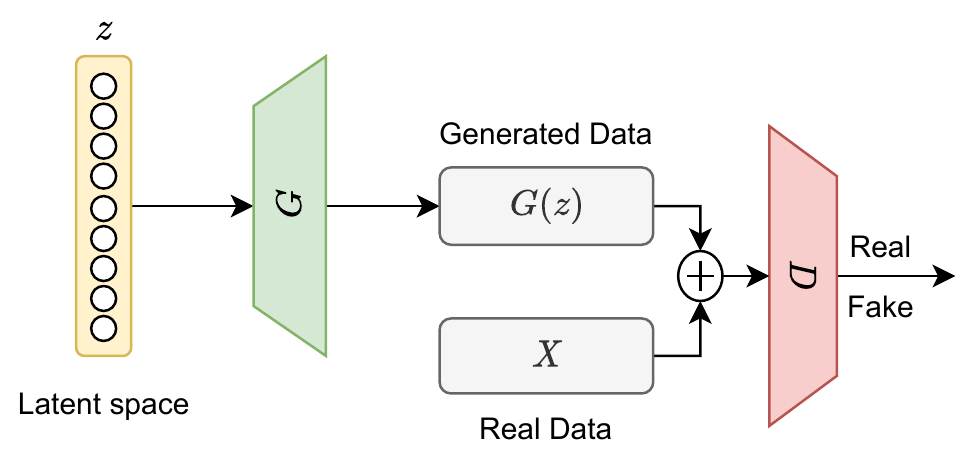}
  \caption{Architecture of the standard GAN.}
  \label{fig:GAN}
\end{subfigure}%
\begin{subfigure}[b]{.48\textwidth}
  \centering
  \includegraphics[width=1\linewidth]{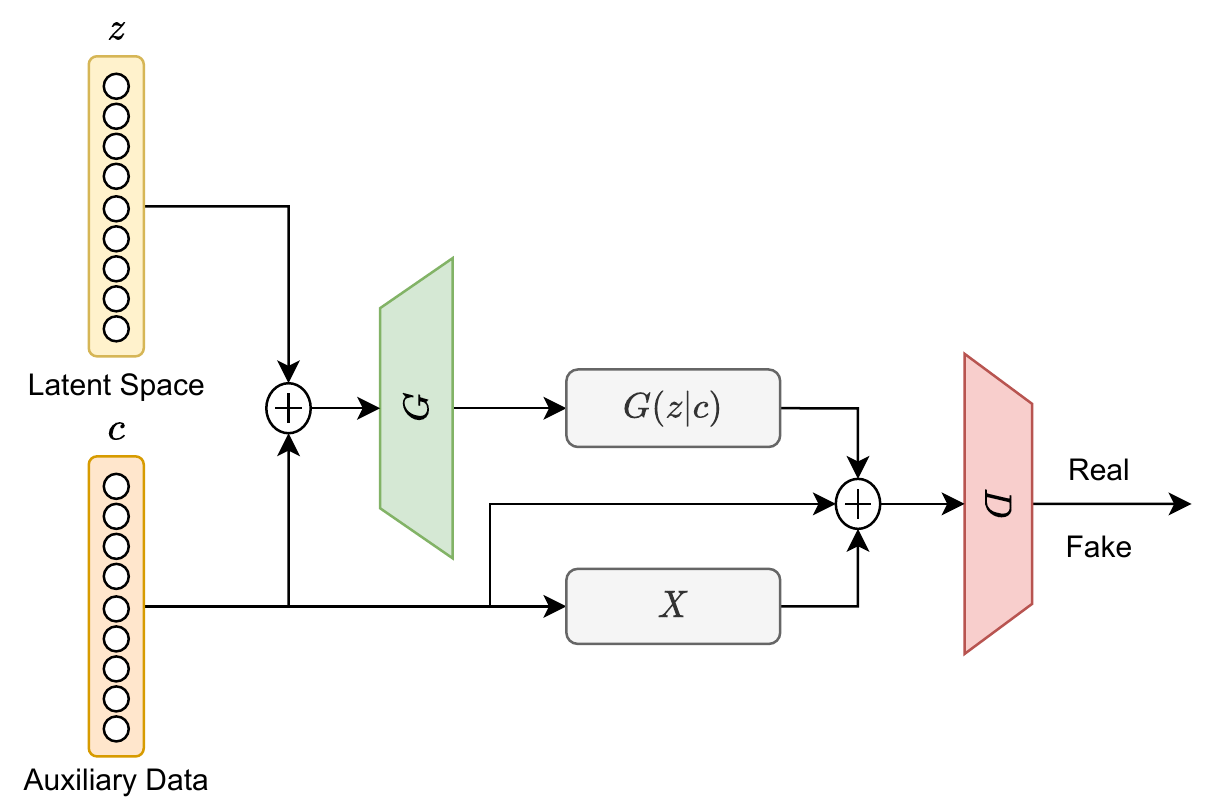}
  \caption{Architecture of the conditional GAN.}
  \label{fig:cGAN}
\end{subfigure}
\begin{subfigure}[b]{.48\textwidth}
  \centering
  \includegraphics[width=1\linewidth]{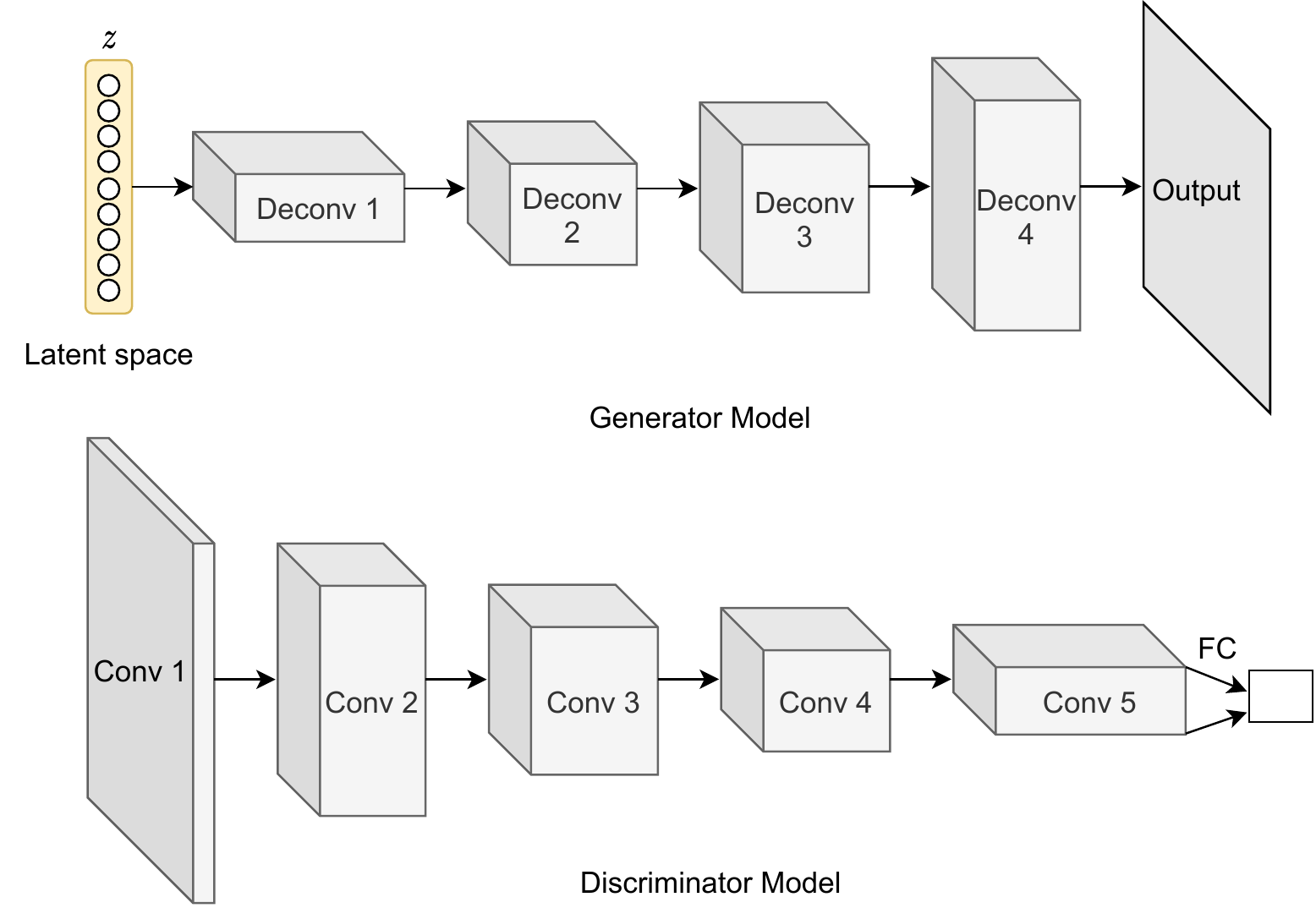}
  \caption{Architecture of the Deep Convolutional GAN.}
  \label{fig:DCGAN}
\end{subfigure}
\begin{subfigure}[b]{.48\textwidth}
  \centering
  \includegraphics[width=1\linewidth]{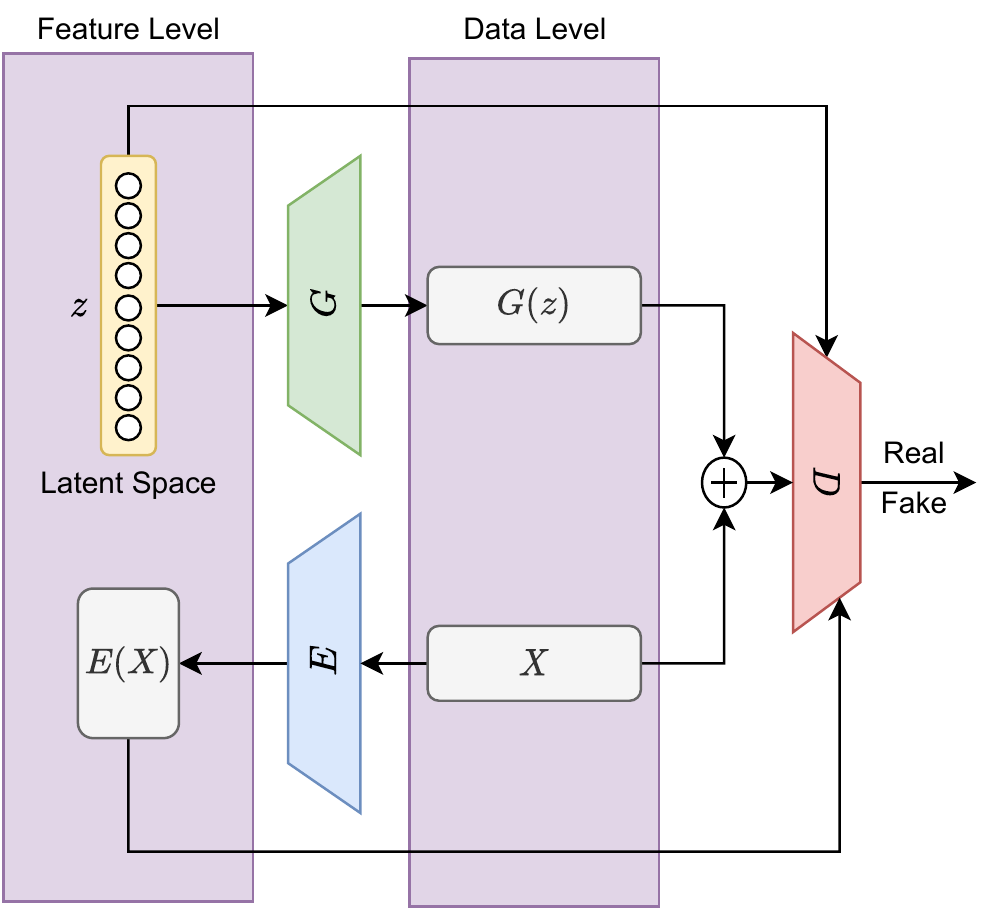}
  \caption{Architecture of the Bi-directional GAN.}
  \label{fig:BiGAN}
\end{subfigure}

\caption{The architectures of the most commonly used generative adversarial networks: (a) the standard GAN, where $G$ and $D$ denote the generator and discriminator; note that the architecture of the Wasserstein GAN is the same; (b)~conditional GAN; the only difference is the addition of auxiliary data, shown as $c$; (c) deep convolutional GAN (DCGAN), where \texttt{Deconv} denotes a deconvolutional layer, \texttt{Conv} a convolutional layer, and \texttt{FC} a fully connected layer; (d) BiGAN, with $E$ denoting the encoder.}
\label{fig:Architectures}
\end{figure*}

The architectures of the top five GAN variants are shown in Figure~\ref{fig:Architectures}. 
We elaborate on the most used architectures, and what makes them suitable for anomaly detection in the following subsections. In addition, we discuss if and how they deal with the challenges mentioned above. 
%PMDONE

% \begin{figure}[htbp!]
% \centering
% \includegraphics[width=1\linewidth]{GAN_Types.pdf}
% \caption{Distribution of different generative adversarial networks used for anomaly detection.}
% \label{fig:GANTypes}
% \end{figure}

\begin{table}[tb]
  \centering
  \small
  \caption{Type of GAN}
    \begin{tabular}{lm{26em}}
    \toprule
    \multicolumn{1}{l}{\textbf{Type of GAN}} & \multicolumn{1}{l}{\textbf{List of papers using this type of GAN}} \\
    \midrule
    
    Deep Convolutional GANs (DCGAN)~\cite{DCGANradford} & ~\cite{ADG11, ADG14, ADG25, ADG26,ADG28, ADG29, ADG31, ADG36, ADG40, ADG46, ADG50, ADG52, ADG58, ADG67, ADG71, ADG74, ADG81, ADG84, ADG90, ADG92, ADG98, ADG100, ADG105, ADG107, ADG110, ADG111, ADG112,ADG113, ADG114, ADG125, ADG126, ADG127, ADG128, ADG129,ADG06, ADG24, ADG65, ADG82, ADG95, ADG116, ADG120} \\    
    Standard GANs~\cite{GANGoodfellow}  & ~\cite{ADG03, ADG05, ADG09, ADG10, ADG13, ADG15, ADG27, ADG32, ADG39, ADG41, ADG42, ADG48, ADG49, ADG51, ADG53, ADG56, ADG57, ADG60, ADG72, ADG79, ADG85, ADG88, ADG94, ADG96, ADG99, ADG108, ADG109, ADG117, ADG118} \\    
    Conditional GANs (cGAN)~\cite{cGANMirza} & ~\cite{ADG01, ADG02, ADG12, ADG18, ADG33, ADG35, ADG54,ADG55, ADG63, ADG64, ADG69, ADG75, ADG77, ADG78, ADG80, ADG91, ADG97, ADG102, ADG106, ADG123, ADG124} \\
    Bi-directional GANs (BiGAN)~\cite{BiGANDonahue} & ~\cite{ADG16, ADG22,ADG30, ADG43, ADG47, ADG61,ADG66, ADG99, ADG103} \\

    % Anomaly GANs (AnoGAN)~\cite{AnoGANSchlegl} & ~\cite{ADG06, ADG24, ADG65, ADG82, ADG95, ADG116, ADG120} \\
    Wasserstein GANs (WGAN)~\cite{WGANArjovsky} & ~\cite{ADG04, ADG21, ADG66, ADG83, ADG87, ADG99, ADG110} \\ 
    Wasserstein GANs with Gradient Penalty (WGAN-GP)~\cite{WGAN-GPGulrajani} & ~\cite{ADG38, ADG62, ADG76, ADG121} \\
    Variational Auetoencoder GANs (VAE-GAN)~\cite{VAE-GANLarsen} & ~\cite{ADG07, ADG08, ADG115, ADG119} \\
    Cycle-GAN~\cite{CycleGANZhu} & ~\cite{ADG19, ADG45, ADG70, ADG86} \\
    \multicolumn{1}{l}{Auxiliary GANs (AC-GAN)~\cite{ACGANOdena}} & ~\cite{ADG17, ADG20, ADG104} \\
    Progressive Growing GANs (PG-GAN)~\cite{PGGANkarras} & ~\cite{ADG17, ADG89} \\
    Orthogonal GAN (O-GAN)~\cite{OGANSu} & ~\cite{ADG23, ADG73} \\
    \multicolumn{1}{l}{Adversarial AutoEncoders (AAE)~\cite{makhzani2015adversarial}} & ~\cite{ADG122} \\
    Balancing GANs (BGAN)~\cite{BAGANMariani} & ~\cite{ADG17} \\
    Energy-Based GANs (EBGAN)~\cite{EBGANZhao} & ~\cite{ADG54} \\
    Dual Discriminator GANs (D2GAN)~\cite{D2GANnguyen} & ~\cite{ADG70} \\
    \multicolumn{1}{l}{GANs with Quadratic Potential (GAN-QP)~\cite{GAN-QPSu}} & ~\cite{ADG73} \\
    One-Class GAN (OCGAN)~\cite{OCGANperera} & ~\cite{ADG101} \\
    Patch GANs (PatchGAN)~\cite{PatchGANisola} & ~\cite{ADG37} \\
    Relativistic Discriminator GANs (RaSGAN)~\cite{RasGANjolicoeur} & ~\cite{ADG93} \\
    Sequence GANs (SeqGAN)~\cite{SeqGANYu} & ~\cite{ADG59} \\
    Text GANs (TextGAN)~\cite{TextGANZhang} & ~\cite{ADG68} \\

    \bottomrule
    \end{tabular}%
  \label{tab:GANTypes}%
\end{table}%"

\begin{table}[tb]
  \centering
  \small
  \caption{Type of GAN used in each application domain.\textbf{AS}: Autonomous Systems, \textbf{CC}: Climate Changes, \textbf{FD}: Fraud Detection, \textbf{HI}: Hyperspectral Images, \textbf{IR}: Image Recognition, \textbf{ID}: Intrusion Detection, \textbf{MA}: Manufacturing, \textbf{ME}: Medical, \textbf{PE}: Power/Energy, \textbf{SS}: Software Systems, \textbf{SU}: Surveillance,\textbf{SH}: System Health, \textbf{TE}: Text, \textbf{TD}: Trajectory Detection, \textbf{VA}: Various.}
    \begin{tabular}{lccccccccccccccc}
    \toprule
    \textbf{Type of GAN} & \textbf{AS}    & \textbf{CC}    & \textbf{FD}    & \textbf{HI}    & \textbf{IR}    & \textbf{ID}    & \textbf{MA}    & \textbf{ME}    & \textbf{PE}    & \textbf{SS}    & \textbf{SU}    & \textbf{SH}    & \textbf{TE}    & \textbf{TD}    & \textbf{VA} \\
    \midrule
    DCGANs & \checkmark & \checkmark & \checkmark &       & \checkmark & \checkmark & \checkmark & \checkmark & \checkmark &       & \checkmark & \checkmark &       &       &  \\
    Standard GANs &       &       & \checkmark & \checkmark &       & \checkmark &       & \checkmark &       & \checkmark & \checkmark & \checkmark &       & \checkmark & \checkmark \\
    BiGANs &       &       &       &       & \checkmark & \checkmark &       &       &       & \checkmark & \checkmark & \checkmark &       & \checkmark & \checkmark \\
    cGANs & \checkmark &       &       &       &       & \checkmark &       & \checkmark &       &       & \checkmark & \checkmark &       &       & \checkmark \\
    WGANs &       &       & \checkmark & \checkmark & \checkmark & \checkmark &       & \checkmark &       &       &       &       &       &       & \checkmark \\
    WGANs-GP &       &       &       &       & \checkmark & \checkmark &       & \checkmark & \checkmark &       &       &       &       &       &  \\
    VAE-GANs &       &       &       &       &       & \checkmark &       & \checkmark &       &       & \checkmark &       &       &       &  \\
    Cycle-GAN &       &       &       &       &       & \checkmark & \checkmark & \checkmark & \checkmark &       &       &       &       &       &  \\
    AC-GANs & \checkmark &       &       & \checkmark &       &       &       &       &       &       &       & \checkmark &       &       &  \\
    PG-GANs &       &       &       &       &       &       &       & \checkmark &       &       &       & \checkmark &       &       &  \\
    O-GANs &       &       &       &       &       &       &       & \checkmark &       &       &       &       &       &       & \checkmark \\
    AAEs &       &       &       &       &       &       &       & \checkmark &       &       &       &       &       &       &  \\
    BGANs &       &       &       &       &       &       &       &       &       &       &       & \checkmark &       &       &  \\
    EBGANs & \checkmark &       &       &       &       &       &       &       &       &       &       &       &       &       &  \\
    D2GANs &       &       &       &       &       &       & \checkmark &       &       &       &       &       &       &       &  \\
    GANs-QP &       &       &       &       &       &       &       &       &       &       &       &       &       &       & \checkmark \\
    OCGANs &       &       &       &       &       &       &       & \checkmark &       &       &       &       &       &       &  \\
    PatchGANs &       &       &       &       &       &       &       &       &       &       & \checkmark &       &       &       &  \\
    RaSGANs &       &       &       &       &       & \checkmark &       &       &       &       &       &       &       &       &  \\
    SeqGANs &       &       &       &       &       & \checkmark &       &       &       &       &       &       &       &       &  \\
    TextGANs &       &       &       &       &       &       &       &       &       &       &       &       & \checkmark &       &  \\
    \bottomrule
    \end{tabular}%
  \label{tab:archvsapp}%
\end{table}%

\subsubsection{Standard Generative Adversarial Networks (GAN)} 

In the first GAN architecture, proposed by Goodfellow et al.~\cite{GANGoodfellow}, the generator and discriminator models are defined by fully connected multilayer perceptrons. For the generator model, to learn the  distribution of the generator $p_{g}$ over data $x$, a prior on input noise variable $p_{z}(z)$ must be defined. This mapping is represented as $G(z;\theta_{g})$, where G is a differentiable function represented by a multilayer perceptron with parameters $\theta_{g}$. For the discriminator model, another multilayer perceptron $D(z;\theta_{d})$ is defined. Its single scalar output represents the probability that $x$ comes from the data rather than $p_{g}$. The training goal for the discriminator model $D$ is to maximize the probability of assigning the correct label to both training examples and samples from the generator model $G$. Simultaneously, $G$ is trained to minimize $\log(1-D(G(z))$. $D$ and $G$ are pitted against each other following a two-player minimax game.
%with value function $V(G,D)$~\cite{GANGoodfellow}:
%\begin{equation} \label{eq:GAN}
%    \min_{G}\max_{D} V(D,G) = \mathbb{E}_{x_\sim p_{data(x)}}[\log D(x)]+\mathbb{E}_{z_\sim p_{z(z)}}[\log(1-D(G(z)))]
%\end{equation}

%Theoretically, based on a two-player minimax game, the generator and the discriminator's training should be arranged in an alternating manner. However, this might lead to overfitting of the discriminator. Consequently, Goodfellow et al. suggest alternating $k$-steps of optimizing $D$ and optimizing $G$. In addition, by modifying equation~\ref{eq:GAN}, one can train $G$ to maximize $\log D(G(z))$ instead of minimizing the $\log (1-D(G(z))$. This provides much stronger gradients early in the learning process. The originally proposed architecture for the generator is a mixture of RELU\cite{RELU} and sigmoid activation functions, and maxout~\cite{MAXOUT} with dropout~\cite{Dropout} for the discriminator.

The Standard GAN optimizes the Jensen-Shannon (JS) divergence to learn the distribution of the data. Consequently, it suffers from an unstable, weak signal when the discriminator is approaching a local optimum, known as the problem of gradient vanishing~\cite{TextGANZhang}. This can also lead to mode collapse. % Several studies attempted to prevent these problems and to achieve better training stability and convergence, e.g., WGAN~\cite{WGANArjovsky} and Least-Squares GAN (LSGAN)~\cite{mao2017least}. 
Another problem of the standard GAN is that it does not provide any inference model to directly capture the inverse mapping. Hence, further training is needed to attain this inference model, adding to the computational cost of the GAN training.
Moreover, as standard GAN uses MLP in the generator and discriminator models, it is not suitable for high dimensional data such as images. This is because MLPs are fully connected networks that require optimization of many parameters. Therefore, more efficient GAN architectures (such as DCGANs) are preferred for images and other high-dimensional data.

\subsubsection{Conditional Generative Adversarial Networks (cGAN)}
Mirza et al. proposed the conditional GAN~\cite{cGANMirza} as an extension to the standard GAN that can control what type of data is generated. For example, a condition can be specified to generate only data of a certain class or type. The conditional model of GAN can be obtained if both the generator and the discriminator are conditioned on some additional information \emph{y} fed through additional input layers. There is no limitation on the type of the data; for example, it can contain class labels or data from different sources~\cite{cGANMirza}. The conditional data generation is advantageous for anomaly detection purposes since cGANs can better generate data from different sources, i.e. multimodal data generation, or it can be used for multimodal anomaly detection. 
%The only difference with the unconditioned counterpart is that we can condition the unconditioned GAN by feeding \emph{y} into both generator and the discriminator as an additional input layer. 
% The adversarial value function for cGAN is defined as follows:

% \begin{equation} \label{eq:cGAN}
%     \min_{G}\max_{D} V(D,G) = \mathbb{E}_{x_\sim p_{data(x)}}[\log D(x|y)]+\mathbb{E}_{z_\sim p_{z(z)}}[\log(1-D(G(z|y)))].
% \end{equation}

% Similar to the standard GAN, this architecture uses ReLu and sigmoid functions in the generator and maxout and dropout in the discriminator. 

% Nevertheless, it still requires careful training and optimization to avoid the problems identified in the standard GAN.

\subsubsection{Deep Convolutional Generative Adversarial Networks (DCGAN)}
Striving to bridge the rift between the success of Convolutional Neural Networks (CNNs) for supervised learning and unsupervised learning, Radford et al.~\cite{DCGANradford} introduced DCGANs, which integrate convolutional neural networks into the standard GAN. DCGANs provide a better network topology for more stable GAN training. The optimization and training processes are the same as for the standard GANs. However, Radford et al. proposed several improvements to the CNNs and Standard GANs. 
% The first one is substituting deterministic pooling functions in the generator and discriminator, e.g., maxpooling, with strided convolutions, i.e. they use all convolutional net~\cite{springenberg2014}. The second modification is to eliminate the use of fully connected layers on top of the convolutional features. The authors claim that removing such layers, e.g., global average pooling, improves the model stability sacrificing the convergence speed. The third change is the use of batch normalization~\cite{ioffe2015}, which improves the stability of the learning process by normalizing the input to each unit to have zero mean and variance of one. Doing so provides better and deeper gradient flow through the network and prevents mode collapse. Nevertheless, applying batch normalization to all layers in the generator and the discriminator leads to some oscillations during the training, which can be avoided by ruling out the generator output layer and input layer of the discriminator from batch normalization. Moreover, they use ReLu for all layers of the generator except for the output layer, which uses Tanh, and for the discriminator, the ReLu is used in all layers.
These modifications are: (1)~using all convolutional nets~\cite{springenberg2014} in the generator and discriminator, (2)~removing fully connected layers on top of the convolutional layer, and (3)~using batch normalization~\cite{ioffe2015}. These changes result in a better model and training stability with deeper gradient flow through the network, preventing mode collapse. %However, Radford et al.~\cite{DCGANradford} show that, if the model is trained too long, a subset of filters will collapse to a single oscillating mode. Therefore, to prevent this issue, one may need to take the training time into consideration.

DCGANs were originally designed for image processing since they employ CNNs. The CNNs allow DCGANs to learn a hierarchy of representations from object parts to scenes in both the generator and discriminator, which makes DCGANs well suited for image anomaly detection.

\subsubsection{Bi-directional Generative Adversarial Networks (BiGAN)} The bi-directional GAN~\cite{BiGANDonahue} adds an autoencoder that learns the mapping of data $x$ to the latent representation $z$ (inference), which makes it well suited for anomaly detection. BiGANs do not make any assumptions about the nature or structure of data. As a result, they provide a general, robust approach for unsupervised representation learning capable of capturing semantic attributes of the data~\cite{BiGANDonahue}. Donahue et al. empirically show that, despite their generality, BiGANs are competitive with the state-of-the-art approaches to perform self-supervised and weakly supervised feature learning tasks. Comparing BiGANs with the standard GAN, the inference mechanism,i.e. feature learning, of BiGANs makes it suitable for anomaly detection techniques since they can be immediately used to generate anomaly scores.

\subsubsection{Wasserstein Generative Adversarial Networks (WGAN)}
In an attempt to alleviate the problem of mode collapse and the challenges of standard GANs to converge to the Nash equilibrium, Arjovsky et al.~\cite{WGANArjovsky} suggested using the Earth-Mover (EM) distance or Wasserstein-1 distance instead of JS divergence used in the standard GAN. Unlike DCGAN, WGAN attempts to enhance the stability of GANs by modifying the adversarial cost function. Arjovsky et al. show that these distances provide gradients that are more useful for updating the generator than the JS divergence function~\cite{WGANArjovsky}.
%Different from standard GAN, where the discriminator model is a binary classifier, the discriminator model in WGAN estimates the Wasserstein distance, therefore the WGAN's discriminator is a regressor.
Although WGAN better handles the problem of mode collapse compared to standard GANs and DCGANs, the weight clipping used in its discriminator made it difficult to converge. Gulrajani et al.~\cite{WGAN-GPGulrajani} proposed an improved version WGAN-GP introducing a gradient penalty to the discriminator model of WGAN instead of weight clipping. This results in better convergence, training speed, and sample quality by forcing the discriminator to learn relatively smoother decision boundaries~\cite{jabbar2020survey}. This improved version of WGAN is already used by several studies for anomaly detection (see Table~\ref{tab:GANTypes}).

\subsection{RQ4: Which type of data instance and datasets are most commonly used for anomaly detection with GANs?}\label{sec:RQ4}
% From the selected studies, we identified six types of input data. As shown in Table~\ref{tab:Datatype}, image and tabular data were mostly used in the primary studies; with a total of 50\% and 26\% portion of the papers, respectively. Meanwhile, text and frequency data were rarely used, only three papers in total utilized these types of data. 
%From the selected studies, we 
We identified six types of input data used for anomaly detection with GANs. As shown in Table~\ref{tab:Datatype}, image is by far the most common type, appearing in 50\% of the examined papers. The two most common application domains for anomaly detection with GANs are related to images: medicine and surveillance. %As a result, images are the most prevailing input data. 
Tabular data is second (26\%), followed by video, time series, text, and frequency data. 

Data preprocessing is a key element that determines the success or failure of many deep learning models~\cite{ADG01, ADG05, ADG37}. We identified 22 types of data preprocessing techniques, summarized in Table~\ref{tab:Pretype}. Owing to images being the most common data type, resizing, normalization, and cropping are the most prevalent preprocessing techniques. These techniques make data more uniform by changing its range and scale. A normalized dataset also speeds up learning. %Therefore, it is essential to apply these techniques in deep learning models. 
Preprocessing is commonly applied to image, tabular and time series data, but rarely to other types of data. It is also worth noting that some studies~\cite{ADG19, ADG80} first transform tabular or frequency data to images before applying other types of preprocessing techniques.

\begin{table}[htbp]
  \centering
  \caption{List of different data types.}
    \begin{tabular}{lcccccc}
    \toprule
    \textbf{Type of input data} & \textbf{Image} & \textbf{Tabular} & \textbf{Video} & \textbf{Time series} & \textbf{Text} & \textbf{Frequency} \\
    \midrule
    \textbf{Number of papers} & 67 & 34 & 19 & 10 & 2 & 1 \\
    \textbf{\% of papers} & 50\% & 26\% & 14\% & 7\% & 2\% & 1\% \\
    \bottomrule
    \end{tabular}%
  \label{tab:Datatype}%
\end{table}%

\begin{table}[htbp]
  \centering
  \caption{List of different preprocessing types with corresponding application to different data types.}
    \begin{tabular}{lccccccc}
    \toprule
    \textbf{Type of preprocessing} & \textbf{\#papers} & \textbf{Image} & \textbf{Tabular} & \textbf{Video} & \textbf{Time series} & \textbf{Text} & \textbf{Frequency} \\
    \midrule
    Resizing & 30 & \checkmark &   & \checkmark &  \checkmark  &  &  \\
    Normalization & 24 &  \checkmark  &  \checkmark  &  &  \checkmark  &  \checkmark  &  \\
    Cropping & 16 &  \checkmark  &  &  &  &  &  \\
    Feature extraction & 12 &  \checkmark  &  \checkmark  &  &  \checkmark  &  &  \\
    Augmentation (\textit{e.g.} flipping) & 8 &  \checkmark  &  \checkmark  &  &  \checkmark  &  &  \\
    Patch extraction & 7 &  \checkmark  &  &  \checkmark  &  &  &  \\
    End-to-end (no preprocessing) & 6 &  \checkmark  &  \checkmark  &  \checkmark  &  \checkmark  &  &  \\
    Frame extraction & 6 &  &  &  \checkmark  &  &  &  \\
    Scaling & 6 &  \checkmark  &  \checkmark  &  &  &  \checkmark  &  \\
    One-hot representation & 5 &  &  \checkmark  &  &  &  &  \\
    Down-sampling & 5 &  \checkmark  &  \checkmark  &  \checkmark  &  &  &  \\
    Data cleaning & 5 &  \checkmark  &  \checkmark  &  &  &  &  \\
    Mapping & 2 &  \checkmark  &  \checkmark  &  &  &  &  \\
    Manually labeling & 2 &  &  \checkmark  &  &  \checkmark  &  &  \\
    Label re-encoding & 2 &  &  \checkmark  &  &  &  &  \\
    Denoising & 2 &  \checkmark  &  &  &  &  &  \\
    Transforming to image & 2 &  &  \checkmark  &  &  &  &  \checkmark  \\
    {Splitting} & 1 &  &  &  &  &  \checkmark  &  \\
    {Edging} & 1 &  \checkmark  &  &  &  &  &  \\
    {Dimension reduction} & 1 &  \checkmark  &  &  &  &  &  \\
    \bottomrule
    \end{tabular}%
  \label{tab:Pretype}%
\end{table}%

 Tables~\ref{tab:dataset1} and~\ref{tab:dataset2} show the datasets used in the primary studies as well as their associated application domains. The ``custom dataset'' stands for a dataset that was either constructed by the study authors or that contains proprietary data not released to the public domain. These tables show that the majority of the utilized datasets are custom, while \emph{UCSD pedestrian}~\cite{mahadevan2010anomaly}, \emph{MNIST}~\cite{lecun2010mnist}, and \emph{CIFAR-10}~\cite{krizhevsky2009learning} are the most commonly used publicly available datasets. 

The \textit{UCSD anomaly detection} dataset was acquired by a stationary camera that captures pedestrian walkways. It includes two subsets: \textit{Ped1} with 34 training and 36 testing video sequences, and \textit{Ped2} with 16 training and 12 testing video sequences. In the normal setting, the video in this dataset contains only pedestrians. Abnormal events occur when either nonpedestrian entities are in the walkway, or there are anomalous pedestrian motion patterns, such as people walking across a walkway or in the grass that surrounds it. This dataset is challenging due to the low-resolution images, different types of moving objects, and the presence of one or more anomalies in the scene. 

The \textit{MNIST} and \textit{CIFAR-10} datasets both appear in 7\% of the examined studies. The \textit{MNIST} database of handwritten digits has a training set of 60,000 examples and a test set of 10,000 examples. The \textit{CIFAR-10} dataset is a collection of 60,000 colour images arranged in 10 object classes of equal size. When MNIST and CIFAT-10 are used in anomaly detection studies, one class is simulated as abnormal and removed from the training class, while the remaining classes are treated as normal. 

The \textit{UMN crowd} dataset~\cite{mehran2009abnormal} is used in 4\% of the examined papers. It contains normal and abnormal crowd behaviour captured in indoor and outdoor scenes of the University of Minnesota. The dataset contains 11 videos with a total of 7,736 frames that were captured under several scenarios at three different indoor and outdoor scenes.  %Scene 1 is an outdoor scene, which contains two scenarios. Scene 2 contains six scenarios, which were captured indoors. Scene 3 contains three scenarios that were captured in an outdoor scene.

\begin{table}[htbp]
\renewcommand\thetable{10-a}
  \small
%   \scriptsize
  \centering
  \caption{The list of identified datasets from primary studies. \textbf{AS}: Autonomous Systems, \textbf{CC}: Climate Changes, \textbf{FD}: Fraud Detection, \textbf{HI}: Hyperspectral Images, \textbf{IR}: Image Recognition, \textbf{ID}: Intrusion Detection, \textbf{MA}: Manufacturing, \textbf{ME}: Medical, \textbf{PE}: Power/Energy, \textbf{SS}: Software Systems, \textbf{SU}: Surveillance,\textbf{SH}: System Health, \textbf{TE}: Text, \textbf{TD}: Trajectory Detection, \textbf{VA}: Various. Note that the datasets highlighted in bold are applied in more than one domain.} 
    \begin{adjustbox}{max width=\columnwidth}
        \begin{tabular}{p{0.45\textwidth}|p{0.02\textwidth}|p{0.02\textwidth}|p{0.02\textwidth}|p{0.02\textwidth}|p{0.02\textwidth}|p{0.02\textwidth}|p{0.02\textwidth}|p{0.02\textwidth}|p{0.02\textwidth}|p{0.02\textwidth}|p{0.02\textwidth}|p{0.02\textwidth}|p{0.02\textwidth}|p{0.02\textwidth}|p{0.02\textwidth}}
        \toprule
        \textbf{Name of Dataset} & AS & CC & FD& HI &IR & ID & MA & ME & PE & SS & SU & SH & TE & TD & VA\\
        \midrule
        20Newsgroups: ~\cite{ADG68} &   & & & & & & & & & & & &\checkmark & &   \\ 
        ABU: ~\cite{ADG32}&   & & &\checkmark & & & & & & & & & & &  \\ 
        ADFA-LD: ~\cite{ADG19, ADG59}&   & & & & & \checkmark& & & & & & & & & \\ 
        ADNI: ~\cite{ADG124} &   & & & & & & &\checkmark & & & & & & & \\ 
        AI city challenge 2019: ~\cite{ADG13} &   & & & & & & & & & & \checkmark& & & & \\
        \textbf{ARRHYTHMIA}: ~\cite{ADG53, ADG61, ADG73} &   & & & & & & & & & & & & & & \checkmark\\
        BratS18: ~\cite{ADG86, ADG124} &   & & & & & & & \checkmark& & & & & & & \\ 
      \textbf{CALTECH-256}: ~\cite{ADG14, ADG85, ADG120} &   & & & &\checkmark & & \checkmark& & & & & & & &\checkmark \\
        CARDIO: ~\cite{ADG03}&   & & & & & & & & & & & \checkmark& & & \\
        Cardiotocography: ~\cite{ADG122}&   & & & & & & & \checkmark& & & & & & &\\
        CCSD-NL: ~\cite{ADG70} &   & & & & & & \checkmark& & & & & & & & \\ 
        CDnet2014: ~\cite{ADG90} &   & & & & & & & & & & \checkmark& & & & \\
        CelebA: ~\cite{ADG101} &   & & & &\checkmark & & & & & & & & & & \\ 
        CICIDS2017: ~\cite{ADG01, ADG92, ADG94, ADG99, ADG106} &   & & & & & \checkmark& & & & & & & & & \\
        \textbf{CIFAR-10}: {~\cite{ADG34, ADG47, ADG48, ADG52, ADG61, ADG62, ADG73, ADG98, ADG101, ADG110, ADG113}} 
        &   & & & &\checkmark & & & & & & & & & & \checkmark\\ 
        CIFAR-100: ~\cite{ADG52} & & & & &\checkmark & & & & & & & & & & \\ 
        COIL-100: ~\cite{ADG62, ADG101} &   & & & & \checkmark& & & & & & & & & & \\ 
        CRACK: ~\cite{ADG29} &   & & & & & &\checkmark & & & & & & & & \\
        Credit card fraud detection: ~\cite{ADG83} &   & &\checkmark & & & & & & & & & & & & \\ 
        CUHK avenue: ~\cite{ADG37, ADG39, ADG64, ADG79} &   & & & & & & & & & &\checkmark & & & &  \\ 
        \textbf{Custom}: {~\cite{ADG05, ADG06, ADG15, ADG17, ADG20, ADG25, ADG26, ADG28, ADG30, ADG31, ADG41, ADG45, ADG48, ADG50, ADG54, ADG55, ADG56, ADG58, ADG71, ADG75, ADG77, ADG80, ADG81, ADG82, ADG84, ADG87, ADG91, ADG95, ADG96, ADG97, ADG104, ADG105, ADG115, ADG116, ADG117, ADG119, ADG121, ADG127, ADG128, ADG129}} &\checkmark & & \checkmark & \checkmark &\checkmark & \checkmark& \checkmark& \checkmark& \checkmark&\checkmark & \checkmark& \checkmark& & & \checkmark \\ 
        CVC-Clinic: ~\cite{ADG69} &   & & & & & & & \checkmark& & & & & & & \\ 
        CVC-ClinicVideoDB: ~\cite{ADG69} &   & & & & & & &\checkmark & & & & & & & \\ 
        CWRU: ~\cite{ADG50} &   & & & & & & & & & & &\checkmark & & & \\
        DAD: ~\cite{ADG12} & \checkmark & & & & & & & & & & & & & & \\
        DDSM: ~\cite{ADG107, ADG112} &   & & & & & & &\checkmark & & & & & & & \\ 
        \textbf{ECG  time\_series}: ~\cite{ADG49} &   & & & & & & & & & & & & & & \checkmark\\ 
        El segundo: ~\cite{ADG21} &   & & & \checkmark & & & & & & & & & & & \\
        Fashion-MNIST: ~\cite{ADG62, ADG101} &   & & & &\checkmark & & & & & & & & & & \\
        Faster R-CNN: ~\cite{ADG46} &   & & & & & & & &\checkmark & & & & & & \\
        \textbf{FFOB}: ~\cite{ADG113}&   & & & & & & & & & & & & & &\checkmark\\
        GEFCom2012: ~\cite{ADG07} &   & & & & &\checkmark & & & & & & & & & \\
        Geolife GPS trajectory: ~\cite{ADG16}&   & & & & & & & & & & & & & \checkmark&\\
        HYDICE: ~\cite{ADG32}&   & & & \checkmark& & & & & & & & & & &\\
        \textbf{IONOSPHERE}: ~\cite{ADG03, ADG53} &   & & & & & & & & & & & \checkmark& & & \checkmark\\
        IRIS: ~\cite{ADG93} &   & & & & &\checkmark & & & & & & & & & \\
        \textbf{IR-MNIST}: ~\cite{ADG118, ADG120} &   & & & & & & \checkmark& & & & & & & & \checkmark\\ 
        Joint european torus: ~\cite{ADG40} &   & & & & & &\checkmark & & & & & & & & \\ 
        \textbf{KDD-Cup99 10\%}: {~\cite{ADG27, ADG34, ADG47, ADG48, ADG61, ADG66, ADG73}} &   & & & &\checkmark& \checkmark& & & & & & & & &\checkmark \\
        LIDC-IDRI: ~\cite{ADG38} &   & & & & & & &\checkmark & & & & & & & \\ 
        LiTS: ~\cite{ADG86} & & & & & & & \checkmark & & & & & & & &\\
        LSUN: ~\cite{ADG101, ADG110} &   & & & &\checkmark & & & & & & & & & &\\ 
        Lymphography\&Mammography: ~\cite{ADG122} &   & & & & & & &\checkmark & & & & & & & \\
        MIT-BIH: ~\cite{ADG24} &   & & & & & & &\checkmark & & & & & & &\\
        \textbf{MNIST}: {~\cite{ADG14, ADG16, ADG40, ADG43, ADG48, ADG62, ADG73, ADG85, ADG98, ADG100, ADG101, ADG110, ADG113, ADG120}} 
        &   & & & & \checkmark& & \checkmark& & & & & & & &\checkmark\\
        MVTec: ~\cite{ADG65} &   & & & &\checkmark & & & & & & & & & &\\
        NIH chest X-ray: ~\cite{ADG57, ADG114} &   & & & & & & & \checkmark& & & & & & & \\
        NIH PLCO: ~\cite{ADG126} &   & & & & & & &\checkmark & & & & & & & \\
        NSL-KDD: ~\cite{ADG01, ADG76, ADG88, ADG92} &   & & & & & \checkmark & & & & & & & & & \\
        \textbf{NYC\_TAXI}: ~\cite{ADG49} &   & & & & & & & & & & & & & &\checkmark \\
        \textbf{PIMA}: ~\cite{ADG02, ADG53} &   & & & & & & & & & & & & & & \checkmark\\
        OCT: ~\cite{ADG23} & & & & & & & \checkmark & & & & & & & &\\
        RSNA: ~\cite{ADG89} &   & & & & & & &\checkmark & & & & & & & \\
        RWC: ~\cite{ADG83} &   & &\checkmark& & & & & & & & & & & & \\
        \bottomrule
        \end{tabular}%
    \end{adjustbox}
  \label{tab:dataset1}%
\end{table}%

\begin{table}[htbp]
 \renewcommand\thetable{10-b}
  \small
%   \scriptsize
  \centering
  \caption{The list of identified datasets from primary studies. \textbf{AS}: Autonomous Systems, \textbf{CC}: Climate Changes, \textbf{FD}: Fraud Detection, \textbf{HI}: Hyperspectral Images, \textbf{IR}: Image Recognition, \textbf{ID}: Intrusion Detection, \textbf{MA}: Manufacturing, \textbf{ME}: Medical, \textbf{PE}: Power/Energy, \textbf{SS}: Software Systems, \textbf{SU}: Surveillance,\textbf{SH}: System Health, \textbf{TE}: Text, \textbf{TD}: Trajectory Detection, \textbf{VA}: Various. Note that the datasets highlighted in bold are applied in more than one domain.}
    \begin{adjustbox}{max width=\columnwidth}
        \begin{tabular}{p{0.45\textwidth}|p{0.02\textwidth}|p{0.02\textwidth}|p{0.02\textwidth}|p{0.02\textwidth}|p{0.02\textwidth}|p{0.02\textwidth}|p{0.02\textwidth}|p{0.02\textwidth}|p{0.02\textwidth}|p{0.02\textwidth}|p{0.02\textwidth}|p{0.02\textwidth}|p{0.02\textwidth}|p{0.02\textwidth}|p{0.02\textwidth}}
        \toprule
        \textbf{Name of Dataset} & AS & CC & FD& HI &IR & ID & MA & ME & PE & SS & SU & SH & TE & TD & VA\\
        \midrule
        San diego: ~\cite{ADG21, ADG32, ADG72} &   & & & \checkmark& & & & & & & & & & & \\
        San Francisco cabspotting: ~\cite{ADG16}&   & & & & \checkmark& & & & & & & & & & \\
        SBHAR: ~\cite{ADG42}&   & & & & &\checkmark & & & & & & & & & \\
        SD-OCT: ~\cite{ADG04} &   & & & & & & & \checkmark& & & & & & & \\
        Sentence polarity: ~\cite{ADG68} &   & & & & & & & & & & & &\checkmark & & \\
        ShanghAaiTech: ~\cite{ADG37, ADG79} &   & & & & & & & & & & \checkmark& & & & \\
        SIXray: ~\cite{ADG103} &   & & & & & & & & & & \checkmark& & & & \\
        Spectralis OCT: ~\cite{ADG18} &   & & & & & & &\checkmark & & & & & & & \\
        SWaT system: ~\cite{ADG108}&   & & & & & \checkmark& & & & & & & & & \\ 
        \textbf{SVHN}: ~\cite{ADG47, ADG61} &   & & & & & & & & & & & & & & \checkmark \\
        \textbf{TalkingData AdTracking}: ~\cite{ADG10, ADG66}&   & &\checkmark & & & & & & & & & & & & \checkmark\\
        Tennessee eastman: ~\cite{ADG03, ADG22} &   & & & & & & & & & & &\checkmark & & & \\
        Texas coast: ~\cite{ADG21}&   & & &\checkmark& & & & & & & & & & &\\
        Thyroid: ~\cite{ADG122} & & & & & & & \checkmark & & & & & & & &\\
        UBA: ~\cite{ADG113} &   & & & & & & & & & & & & & &\checkmark\\
        \textbf{UCI}: ~\cite{ADG34, ADG83} &  & & \checkmark & & & & & & & & & & & & \checkmark \\
        \textbf{UCSD}:  {~\cite{ADG09, ADG11, ADG33, ADG35, ADG37, ADG51, ADG63, ADG64, ADG67, ADG78, ADG79, ADG85, ADG102, ADG118, ADG125}} &   & & & & & & & & & & \checkmark& & & &\checkmark\\
        \textbf{Udacity}: ~\cite{ADG54, ADG60} &   & & & & & & & & & & & & & \checkmark& \\
        \textbf{UMN}: {~\cite{ADG35, ADG39, ADG63, ADG64, ADG78, ADG102, ADG118}} &   & & & & & & & & & &\checkmark & & & &\checkmark\\
        UNSW-NB15: ~\cite{ADG01} &   & & & & &\checkmark & & & & & & & & &\\
        VIRAT: ~\cite{ADG90} &   & & & & & & & & & & \checkmark & & & &\\
        WADI test-bed: ~\cite{ADG108} &   & & & & &\checkmark & & & & & & & & & \\
        WOA13 monthly: ~\cite{ADG111} &   &\checkmark & & & & & & & & & & & & &\\
        WOOD: ~\cite{ADG29} &   && & & & &\checkmark & & & & & & & &\\
        \bottomrule
        \end{tabular}%
    \end{adjustbox}
  \label{tab:dataset2}%
\end{table}%

% UCI datasets include KDD-Cup99 10%,  cover type-A, cover type-B and shuttle
From the perspective of application domain, we found that studies on fraud detection use the \textit{Credit Card Fraud Detection} dataset, \textit{real world credit (RWC) dataset}, \textit{TalkingData AdTracking}, \textit{UCI dataset} and a custom dataset. For surveillance anomaly detection purposes, 74\% studies use the \textit{CUHK avenue}, \textit{ShanghAaiTech}, \textit{UCSD}, and \textit{UMN} datasets. Among all datasets identified in the primary studies, twelve are used for intrusion detection, seven for manufacturing anomaly detection, fifteen for medical anomaly detection, and twelve for image anomaly detection. While most datasets are not used across all domains, eighteen are used in multiple domains as highlighted in Table~\ref{tab:dataset1} and Table~\ref{tab:dataset2}. For instance, \textit{KDD-cup99 10\%} is used in both image recognition~\cite{ADG47} and intrusion detection~\cite{ADG48} while \textit{MNIST} is used in image recognition~\cite{ADG16, ADG40, ADG43, ADG48, ADG62,ADG98, ADG113, ADG120} and manufacturing anomaly detection~\cite{ADG73}.

\subsection{RQ5: Which metrics are used to evaluate the performance of GANs in generating data and anomaly detection?}\label{sec:RQ5}

%Table~\ref{tab:TableTask} shows that 34 papers applied data augmentation using GANs while 94 papers implemented representation learning using GANs. 
We found that 27 out of 128 primary studies evaluated the quality of the generated samples %. To measure the quality of the generated data, some papers adopted a specific metric to evaluate the similarity between the original sample and the generated sample. We identified 
using 9 different performance evaluation metrics. Most studies evaluated data quality quantitatively, while six papers implemented visual inspection to evaluate the quality of the generated samples~\cite{ADG04,ADG36,ADG58,ADG71,ADG91,ADG126}. During the inspections, the generated samples were examined by application domain experts, or simply the authors of the individual studies. Quantitative evaluation was performed using eight performance metrics, most commonly the structural similarity index measure (SSIM) and peak signal-to-noise ratio (PSNR) that were each used in 26\% of the studies that evaluated performance.

SSIM, adopted by seven papers ~\cite{ADG18,ADG52,ADG54,ADG62, ADG64, ADG115, ADG116}, quantifies the relative perceptual similarity between two images. This metric ranges from -1 to 1, with 1 indicating a perfect pixel match between the original and generated samples, -1 corresponding to inverted images, and 0 marking no similarity~\cite{wang2004image}. Seven papers ~\cite{ADG09,ADG11,ADG13,ADG37,ADG40,ADG86, ADG97} used PSNR as a metric to measure the quality of the generated images. This metric evaluates the similarity of two samples through the ratio of the total number of pixels divided and the mean squared error between the original and generated images. A higher value of the PSNR indicates that the generated sample is closer to the original. 
%
%
% To calculate SSIM, Eq. (\ref{eqn:SSIM}) can be applied, where $C_n = (K_nL)^2$, $K_n\ll1$ is a small constant and L is the dynamic range of the pixel values (255 for 8-bit grayscale images).
%
% \begin{equation}
% \label{eqn:SSIM}
% SSIM(x,y) = \frac{(2\mu_x\mu_y+C_1)(2\sigma_{xy}+C_2)}{(\mu_x^2+\mu_y^2+C_1)(\sigma_{x}^2+\sigma_{y}^2+C_2)}
% \end{equation}
%
% \begin{equation}
% \label{eqn:PSNR}
%     PSNR(\widehat{I},I) = 10* \log_{10} (\frac{[max{\widehat{I_i}}]^2}{\frac{1}{N}\sum_{i=0}^{N} (I_i-\widehat{I_i})^2 })
% \end{equation}
%
The Fréchet inception distance (FID), adopted by two papers~\cite{ADG70,ADG93}, is a widely used evaluation method for evaluating the diversity and similarity of generated images~\cite{heusel2017gans}. By calculating and comparing the feature vectors of a collection of real and generated images, FID can measure the distance between the real and generated distribution.

% Earth mover’s distance (EMD) was introduced by ~\cite{rubner2000earth} and was implemented in ~\cite{ADG09} as a evaluation criteria to assess the quality of the generated frames as well as to decide whether a frame is anomalous. 

% Moreover, Eq. (\ref{eqn:PSNR}) shown how to calcualte for FID, where $\mu_r$ and $\mu_g$ represent the means of the real image features and the generated image features, respectively, and $C_r$ and $C_g$ are the covariance values of the real image features and the generated image features, respectively.
% \begin{equation}
% \label{eqn:FID}
%     FID(P_r,Pg) = \|\mu_r+\mu_g\| + Tr(C_r+C_g-2(C_rC_g))^{1/2}
% \end{equation}

% To calculate EMD, consider two distributions P and Q with m and n clusters, respectively. Let $d_{i,j}$ represents the ground distance between two clusters $p_i$ and $q_i$; $f_{i,j}$ represents the flow between two clusters $p_i$ and $q_i$, that minimize the overall cost. 

% \begin{equation}
% \label{eqn:EMD}
%     EMD(P,Q) = \frac{\sum_{i=1}^{m}\sum_{j=1}^{n}f_{i,j}d_{i,j}}{\sum_{i=1}^{m}\sum_{j=1}^{n}f_{i,j}}
% \end{equation}

\begin{table}[tbp]
\renewcommand\thetable{11}
  \small
  \centering
  \caption{The performance metrics that are used to evaluate generated samples with correspondent input types. Note that there are no performance metrics used for evaluating tabular, text and frequency data.}
    \begin{tabular}{p{0.45\textwidth}p{0.05\textwidth}p{0.05\textwidth}p{0.05\textwidth}p{0.09\textwidth}p{0.05\textwidth}p{0.07\textwidth}}
    \toprule
    \textbf{Type of performance metrics used} & \textbf{Image} & \textbf{Tabular} &\textbf{Video} &\textbf{Time series} &\textbf{Text} &\textbf{Frequency} \\
    \midrule
    Structural similarity indices metrics (SSIM): ~\cite{ADG17, ADG52, ADG54, ADG62, ADG64, ADG115, ADG116} & \checkmark & & \checkmark\\
    Peak signal to noise ratio (PSNR): ~\cite{ADG09, ADG11, ADG13, ADG37, ADG40, ADG86, ADG97}& \checkmark & & \checkmark  \\
    Visual inspection: ~\cite{ADG04, ADG36, ADG58, ADG71, ADG91, ADG126} & \checkmark & & &\checkmark\\
    Fréchet inception distance (FID): ~\cite{ADG70, ADG93} & \checkmark\\
    Signal to noise ratio (SNR): ~\cite{ADG50} &  & & &\checkmark\\
    L2-norm distance: ~\cite{ADG125}& & & \checkmark \\
    Fully convolutional network (FCN)-score: ~\cite{ADG64} & & & \checkmark\\
    Earth mover's distance: ~\cite{ADG09} & & & \checkmark\\
    Cosine similarity: ~\cite{ADG125} & & & \checkmark\\
    \bottomrule
    \end{tabular}%
  \label{tab:Metric1}%
\end{table}%

We also studied whether performance metrics were used with specific input data types. As shown in Table~\ref{tab:Metric1}, we observed that most performance metrics have been applied to image and video data, while only two papers utilized metrics for time series data ~\cite{ADG50, ADG58}. It is also worth mentioning that frame data is usually extracted from the video before being fed into the GAN. Therefore, the performance metrics are essentially used only to evaluate the image data. We also examined the relationship between the performance metrics and application domains. Table~\ref{tab:Metric2} shows that most metrics were adopted in surveillance anomaly detection to evaluate the generated samples, while most domains such as fraud detection, power/energy anomaly detection, and software systems did not evaluate the quality of the generated samples at all. In addition, SSIM, PSNR, FID, and visual inspection were used in various domains, while other metrics were only applied to one specific domain.

\begin{table}[tbp]
\renewcommand\thetable{12}
  \small
  \centering
  \caption{The performance metrics that are used to evaluate generated samples with correspondent application domains. \textbf{AS}: Autonomous Systems, \textbf{CC}: Climate Changes, \textbf{FD}: Fraud Detection, \textbf{HI}: Hyperspectral Images, \textbf{IR}: Image Recognition, \textbf{ID}: Intrusion Detection, \textbf{MA}: Manufacturing, \textbf{ME}: Medical, \textbf{PE}: Power/Energy, \textbf{SS}: Software Systems, \textbf{SU}: Surveillance,\textbf{SH}: System Health, \textbf{TE}: Text, \textbf{TD}: Trajectory Detection, \textbf{VA}: Various.}
    \begin{tabular}{p{0.33\textwidth}p{0.02\textwidth}p{0.02\textwidth}p{0.02\textwidth}p{0.02\textwidth}p{0.02\textwidth}p{0.02\textwidth}p{0.02\textwidth}p{0.02\textwidth}p{0.02\textwidth}p{0.02\textwidth}p{0.02\textwidth}p{0.02\textwidth}p{0.02\textwidth}p{0.02\textwidth}p{0.02\textwidth}}
    \toprule
    \textbf{Type of Performance Metrics Used} & AS & CC & FD& HI &IR & ID & MA & ME & PE & SS & SU & SH & TE & TD & VA \\
    \midrule
    Structural similarity indices (SSIM): ~\cite{ADG17, ADG52, ADG54, ADG62, ADG64, ADG115, ADG116} 
    & \checkmark & & & &\checkmark & &\checkmark &\checkmark & & & \checkmark&\checkmark & \\
    Peak signal to noise ratio (PSNR): ~\cite{ADG09, ADG11, ADG13, ADG37, ADG40, ADG86, ADG97}
    &  & & & & & & & \checkmark& & & \checkmark& &  \\
    Visual inspection: ~\cite{ADG04, ADG36, ADG58, ADG71, ADG91, ADG126}  
    & \checkmark & & & & & &\checkmark &\checkmark & & & & \checkmark&\\
    Fréchet inception distance (FID): ~\cite{ADG70, ADG93} 
    &  & & & & & \checkmark&\checkmark & & & & & &\\
    Signal to noise ratio (SNR): ~\cite{ADG50}&  & & & & & & & & & & & \checkmark&\\
    L2-norm distance: ~\cite{ADG125}&  & & & & & & & & & &\checkmark & & \\
    Fully convolutional network (FCN)-score: ~\cite{ADG64} &  & & & & & & & & & &\checkmark & &\\
    Earth mover’s distance (EMD): ~\cite{ADG09} &  & & & & & & & & & &\checkmark & &\\
    Cosine similarity: ~\cite{ADG125} &  & & & & & & & & & &\checkmark & &\\
    \bottomrule
    \end{tabular}%
  \label{tab:Metric2}%
\end{table}%

%\subsubsection{Performance Metrics For Anomaly Detection}
%In this section, we study the most used performance metrics for anomaly detection using GANs. According to our study,
The actual anomaly detection performance in the studies is evaluated using different (usually more traditional) metrics.
The area under the receiver operating characteristic curve (AUROC) is the most frequently used metric to evaluate anomaly detection (in 68 primary studies). Precision, F1-score, accuracy, recall, sensitivity, equal error rate (EER), specificity, and receiver operating curve (ROC) are also frequently used as metrics in this area. 
\subsection{RQ6: Which anomaly detection techniques are used along with GANs?}\label{sec:RQ6}
This section discusses the different types of anomaly detection techniques that either use GANs or are compared with GANs. Based on the labelled data availability, anomaly detection techniques are divided into three classes: supervised, semi-supervised, and unsupervised anomaly detection. During data synthesis for RQ6, we noticed that not all primary studies use consistent definitions for these classes. Therefore, we use Chandola et al.'s~\cite{anomalysurvey} definition of supervised, semi-supervised, and unsupervised anomaly detection. In addition, as there is a wide variety of anomaly detection techniques, we only considered the techniques used in more than three primary studies.

% \begin{figure}[htbp]
% \centering
% \includegraphics[width=0.8\linewidth]{Techs.png}
% \caption{Classification of anomaly detection techniques.}
% \label{fig:Techs}
% \end{figure}

\subsubsection{Supervised Anomaly Detection}
Anomaly detection techniques trained in a supervised fashion assume that labelled data for the normal and anomaly classes is available. A training dataset is used to train the model for predicting the class labels, and then the predictive model is evaluated on an unseen test dataset.
Dependence of the supervised anomaly detection techniques on the data labels makes them more vulnerable to the problem of an imbalanced dataset. Therefore, these anomaly detection techniques are biased toward classifying the majority class. 

In 26.3\% of the investigated primary studies, GANs are applied to address the problem of the imbalanced dataset for anomaly detection by augmenting it with the data generated by GANs. 
The list of primary studies that use GANs for data augmentation is shown in Table~\ref{tab:TableTask}. 
The list of papers that used GANs along with supervised anomaly detection techniques, i.e. GAN-assisted approaches, is shown in Table~\ref{tab:supervised}.

% % Table generated by Excel2LaTeX from sheet 'Sheet1'
% \begin{table}[htbp]
% \renewcommand\thetable{13}
%   \centering
%   \caption{GAN-assisted supervised anomaly detection techniques. }
%     \begin{tabular}{lllm{14em}}
%     \toprule
%         \multicolumn{3}{l}{\textbf{Techniques}} & \textbf{References} \\
%     \midrule
%     \multirow{4}[6]{*}{Classification-based} & \multicolumn{2}{l}{Support Vector Machines} & \cite{ADG01,ADG10,ADG15,ADG20,ADG27,ADG28,ADG74,ADG83,ADG84,ADG92,ADG125} \\\cmidrule{2-4}           
%           & \multirow{2}[0]{*}{NN-based methods} & Convolutional Neural Networks & ~\cite{ADG01,ADG70,ADG74,ADG76,ADG91,ADG92,ADG103,ADG107,ADG112,ADG124} \\
%           &       & Multilayer perceptron & ~\cite{ADG01,ADG10,ADG19,ADG59,ADG74,ADG76,ADG104,ADG125} \\
%           \cmidrule{2-4}          
%           & Nearest neighbours & K-Nearest Neighbors & ~\cite{ADG15,ADG27,ADG28,ADG59,ADG125} \\
%     \midrule
%     \multirow{4}[6]{*}{Regression-based } & \multicolumn{2}{l}{Naive Bayes } & ~\cite{ADG01,ADG27,ADG59,ADG92} \\\cmidrule{2-4}          
%           & Ensemble methods & Random forest & ~\cite{ADG01,ADG10,ADG15,ADG27,ADG28,ADG59,ADG92,ADG106,ADG125} \\\cmidrule{2-4}          
%           & Linear model & Logistic regression & ~\cite{ADG10,ADG15,ADG27,ADG59,ADG59,ADG83} \\\cmidrule{2-4}          
%           & \multicolumn{2}{l}{Decision trees} & ~\cite{ADG01,ADG15,ADG27,ADG28,ADG59,ADG125} \\
%     \bottomrule
%     \end{tabular}%
%   \label{tab:supervised}%
% \end{table}%

% Table generated by Excel2LaTeX from sheet 'Sheet2'
\begin{table}[tbp]
\renewcommand\thetable{13}
  \centering
  \caption{GAN-assisted supervised anomaly detection techniques.}
    \begin{tabular}{ll}
    \toprule
    \textbf{Techniques} & \textbf{References} \\
    \midrule
    Support vector machines & ~\cite{ADG01,ADG10,ADG15,ADG20,ADG27,ADG28,ADG74,ADG83,ADG84,ADG92,ADG125} \\
    Neural network-based methods & ~\cite{ADG01,ADG10,ADG19,ADG59,ADG70,ADG74,ADG76,ADG91,ADG92,ADG103,ADG104,ADG107,ADG112,ADG124,ADG125} \\
    Nearest neighbors & ~\cite{ADG15,ADG27,ADG28,ADG59,ADG125} \\

    Naive Bayes  & ~\cite{ADG01,ADG27,ADG59,ADG92} \\
    Ensemble methods & ~\cite{ADG01,ADG10,ADG15,ADG27,ADG28,ADG59,ADG92,ADG106,ADG125} \\
    Linear model & ~\cite{ADG10,ADG15,ADG27,ADG59,ADG59,ADG83} \\
    Decision trees & ~\cite{ADG01,ADG15,ADG27,ADG28,ADG59,ADG125} \\
    \bottomrule
    \end{tabular}%
  \label{tab:supervised}%
\end{table}%

\subsubsection{Semi-Supervised Anomaly Detection}
Anomaly detection techniques trained in a semi-supervised manner assume that the labelled data is available only for the normal class. The main benefit of semi-supervised anomaly detection techniques is that they do not require data for the anomalous class. In the reviewed primary studies, GANs are mostly used in a semi-supervised manner. By training a GAN to learn the distribution of the normal class, a deviation from the normal distribution is identified using an anomaly scoring technique. The list of GAN-based semi-supervised anomaly detection techniques is shown in Table~\ref{tab:SemiGAN}. From all GAN-based techniques, AnoGAN~\cite{ADG128} is the GAN-based technique most often used as a baseline for comparison with newly proposed methods. There are several other techniques that can be used for anomaly detection purposes in a semi-supervised manner, as listed in Table~\ref{tab:semisupervisedtech}. 
% Table generated by Excel2LaTeX from sheet 'Semi'
\begin{table}[tbp]
\renewcommand\thetable{14}
  \centering
  \small
  \caption{Semi-supervised GAN-based anomaly detection techniques.}
    \begin{tabular}{lm{18em}lm{18em}}
    \toprule
    \textbf{Type of GAN} &  \textbf{List of references} & \textbf{Type of GAN} &  \textbf{List of references} \\
    \midrule
    AC-GAN & ~\cite{ADG20} &PatchGAN & ~\cite{ADG37} \\
    BiGAN & ~\cite{ADG22,ADG30,ADG47,ADG61,ADG66,ADG99} & RaSGAN & ~\cite{ADG93} \\
    cGAN  & ~\cite{ADG02,ADG18,ADG33,ADG35,ADG54,ADG55,ADG63,ADG64,ADG75,ADG77,ADG78,ADG80,ADG97,ADG102} & Standard GAN & ~\cite{ADG13,ADG34,ADG39,ADG41,ADG42,ADG48,ADG49,ADG51,ADG56,ADG57,ADG60,ADG79,ADG85,ADG88,ADG94,ADG96,ADG99,ADG108,ADG109,ADG117,ADG118} \\
    Cycle-GAN & ~\cite{ADG45} & TextGAN & ~\cite{ADG68} \\
    DCGAN & ~\cite{ADG11,ADG25,ADG26,ADG29,ADG31,ADG36,ADG40,ADG46,ADG50,ADG52,ADG58,ADG71,ADG81,ADG90,ADG98,ADG100,ADG105,ADG110,ADG113,ADG114,ADG127,ADG128,ADG129,ADG24,ADG65,ADG82,ADG95,ADG116,ADG120} & VAE-GAN &~\cite{ADG115} \\
    EBGAN & ~\cite{ADG54} & WGAN & ~\cite{ADG21,ADG66,ADG87,ADG99,ADG110} \\
    GAN-QP & ~\cite{ADG73} & WGAN-GP & ~\cite{ADG38,ADG62} \\
    O-GAN & ~\cite{ADG23,ADG73} &       &  \\
    \bottomrule
    \end{tabular}%
  \label{tab:SemiGAN}%
\end{table}%

The table shows that several papers used Mixture of Dynamic Texture (MDT)~\cite{li2013anomaly} as an anomaly detection technique in crowded scenes. Mehran et al.~\cite{mehran2009abnormal} use a  generative probabilistic model called Social Force for semi-supervised anomaly detection. Two primary studies investigate sparse dictionary learning-based anomaly detection techniques, detection at 150 FPS~\cite{lu2013abnormal} and sparse reconstruction~\cite{cong2011sparse}. 
% Autoencoder-based anomaly detection techniques are suitable for dimensionality reduction, and they can be also used for extracting features from data. %Recently, they attracted a large number of researchers. 
In the primary studies, the performance of different flavours of autoencoders (AEs) were compared to the performance of GANs in anomaly detection such as standard AEs~\cite{sakurada2014anomaly}, Variational AEs (VAEs),  convolutional AEs (CVAEs)~\cite{an2015variational}, Denoising AEs (DAEs)~\cite{xu2015learning}, and Adversarial AEs (AAEs)~\cite{makhzani2015adversarial}. Moreover, some of the primary studies compared the proposed anomaly detection techniques with a Long Short Term Memory (LSTM) based approach~\cite{du2017deeplog}.

% % Table generated by Excel2LaTeX from sheet 'List of papers'
% \begin{table}[htbp]
% \renewcommand\thetable{15}
%   \centering
%   \caption{Semi-supervised anomaly detection techniques compared to GANs.}
%     \begin{tabular}{llm{17em}}
%     \toprule
%         \multicolumn{2}{l}{\textbf{Techniques}} & \textbf{List of references} \\
%     \midrule
    
%     Dynamic texture & Mixture of dynamic texture & ~\cite{ADG11,ADG35,ADG37,ADG63,ADG64,ADG78,ADG85,ADG102,ADG118} \\\cmidrule{1-3}
%     {Generative probabilistic model} & Social force & ~\cite{ADG35,ADG51,ADG63,ADG64,ADG78,ADG79,ADG102} \\\cmidrule{1-3}
%     \multirow{2}[2]{*}{Sparse dictionary learning} & Detection at 150 FPS & ~\cite{ADG35,ADG63,ADG64,ADG78,ADG79,ADG102} \\
%           & Sparse reconstruction & ~\cite{ADG35,ADG63,ADG64,ADG78,ADG102} \\\cmidrule{1-3}
%     \multirow{5}[2]{*}{Autoencoder-based} & Variational autoencoder & ~\cite{ADG34,ADG66,ADG87,ADG95,ADG100,ADG110,ADG113,ADG115} \\
%           & Convolutional autoencoder & ~\cite{ADG09,ADG11,ADG37,ADG39,ADG51,ADG63,ADG64,ADG79} \\
%           & Standard autoencoder & ~\cite{ADG03,ADG04,ADG18,ADG34,ADG60,ADG64,ADG88,ADG99,ADG108} \\
%           & AMDN (denoising autoencoder) & ~\cite{ADG35,ADG51,ADG63,ADG64,ADG78,ADG85,ADG102} \\
%           & Adversarial autoencoder & ~\cite{ADG04,ADG21,ADG81} \\\cmidrule{1-3}
%     Recurrent neural networks & LSTM  & ~\cite{ADG88,ADG94,ADG95,ADG105} \\
% \midrule    \end{tabular}%
%   \label{tab:semisupervisedtech}%
% \end{table}%

\begin{table}[tbp]
\renewcommand\thetable{15}
  \centering
  \caption{Semi-supervised anomaly detection techniques compared to GANs.}
    \begin{tabular}{lm{32em}}
    \toprule
    \textbf{Techniques} & \textbf{List of references} \\
    \midrule
    Dynamic texture & ~\cite{ADG11,ADG35,ADG37,ADG63,ADG64,ADG78,ADG85,ADG102,ADG118} \\
    Generative probabilistic model & ~\cite{ADG35,ADG51,ADG63,ADG64,ADG78,ADG79,ADG102} \\
    Sparse dictionary learning & ~\cite{ADG35,ADG63,ADG64,ADG78,ADG79,ADG102} \\
    Autoencoder-based & ~\cite{ADG03,ADG04,ADG09,ADG11,ADG18,ADG21,ADG34,ADG35,ADG37,ADG39,ADG51,ADG60,ADG63,ADG64,ADG66,ADG78,ADG79,ADG81,ADG85,ADG87,ADG88,ADG95,ADG99,ADG100,ADG102,ADG108,ADG110,ADG113,ADG115} \\
    Recurrent neural networks & ~\cite{ADG88,ADG94,ADG95,ADG105} \\
    \bottomrule
    \end{tabular}%
  \label{tab:semisupervisedtech}%
\end{table}%

\subsubsection{Unsupervised Anomaly Detection} These types of anomaly detection techniques do not require a labelled dataset. This is based on the central assumption that normal instances are far more frequent than anomalies in the test data~\cite{anomalysurvey}. However, if this assumption is not valid, the anomaly detection will significantly suffer from false alarms. Assuming that the unlabeled dataset contains very few anomalous instances and the model is robust against these few anomalies, we can adapt a semi-supervised anomaly detection technique to work in an unsupervised manner by training the model on a portion of the unlabeled dataset. A list of GAN-based unsupervised anomaly detection techniques is presented in Table~\ref{tab:unsuperGAN}. In addition, Table~\ref{tab:unsupervisedtechs} presents a list of unsupervised anomaly detection techniques that have been considered for anomaly detection and compared to GANs in the literature.

% Table generated by Excel2LaTeX from sheet 'Un'
\begin{table}[tbp]
\renewcommand\thetable{16}
  \centering
  \caption{Unsupervised anomaly detection techniques based on GANs.}
    \begin{tabular}{llll}
    \toprule
    \textbf{GAN architecture} &  \textbf{List of references} & \multicolumn{1}{l}{\textbf{GAN architecture}} & \multicolumn{1}{l}{ \textbf{List of references}} \\
    \midrule
    AAE   & ~\cite{ADG122} & Standard GAN & ~\cite{ADG03,ADG05,ADG09,ADG32,ADG53,ADG72} \\
    BiGAN & ~\cite{ADG16,ADG43} & VAE-GAN & ~\cite{ADG07,ADG08,ADG119} \\
    cGAN  & ~\cite{ADG12} & WGAN  & ~\cite{ADG04,ADG110} \\
    DCGAN & ~\cite{ADG14,ADG110,ADG06} & WGAN-GP & ~\cite{ADG121} \\

    \bottomrule
    \end{tabular}%
  \label{tab:unsuperGAN}%
\end{table}%

From Table~\ref{tab:unsupervisedtechs}, we can observe that one-class classifiers~\cite{scholkopf1999support} have been of great interest from an unsupervised anomaly detection perspective. Isolation forest~\cite{liu2008isolation} is another unsupervised technique competing with GANs in this area. Several variants of principal component analysis ~\cite{tipping1999probabilistic} have also been compared often to GANs in terms of performance. %Most common PCA variants include Dual Principal Component Pursuit (DPCP)~\cite{tsakiris2018dual}, Coherence Pursuit (CoP)~\cite{rahmani2017coherence}, Mixture of Probabilistic Principal Component Analysers (MPPCA)~\cite{li2013anomaly}. 
Several linear models have also been applied, namely REAPER~\cite{lerman2015robust} and Low Rank Representation~\cite{liu2010robust}. Other techniques compared to GANs for anomaly detection include Gaussian mixture models~\cite{yang2009outlier}, R-graph~\cite{you2017provable}, local outlier factor~\cite{breunig2000lof}, deep structured energy-based models ~\cite{zhai2016deep}, and deep autoencoding Gaussian mixture models~\cite{zong2018deep}.

% % Table generated by Excel2LaTeX from sheet 'Sheet3'
% \begin{table}[htbp]
% \renewcommand\thetable{17}
%   \centering
%   \caption{Unsupervised anomaly detection techniques compared to GANs}
%     \begin{tabular}{llm{23em}}
%      \toprule
%         \multicolumn{2}{l}{\textbf{Techniques}} & \textbf{List of references} \\
%     \midrule
%     One-class support vector machines &  & ~\cite{ADG03,ADG05,ADG30,ADG34,ADG36,ADG47,ADG48,ADG49,ADG52,ADG53,ADG60,ADG61,ADG71,ADG73,ADG99,ADG101,ADG108,ADG110,ADG115,ADG127} \\
%     Ensemble methods & Isolation forest & ~\cite{ADG03,ADG05,ADG34,ADG47,ADG49,ADG60,ADG61,ADG109,ADG110,ADG122} \\
%     \multirow{4}[2]{*}{Principal component analysis} & PCA   & ~\cite{ADG22,ADG81,ADG108,ADG109} \\
%           & DPCP  & ~\cite{ADG14,ADG62,ADG85,ADG120} \\
%           & CoP   & ~\cite{ADG14,ADG62,ADG85,ADG120} \\
%           & MPPCA & ~\cite{ADG11,ADG35,ADG37,ADG51,ADG63,ADG64,ADG78,ADG79,ADG85,ADG102} \\
%     \multirow{2}[2]{*}{Linear model} & REAPER & ~\cite{ADG14,ADG62,ADG85,ADG120} \\
%           & LRR   & ~\cite{ADG14,ADG62,ADG85,ADG120} \\
%     Probabilistic model & GMM   & ~\cite{ADG20,ADG36,ADG53,ADG65,ADG110} \\
%     Stochastic model & R-Graph & ~\cite{ADG62,ADG85,ADG120} \\
%     Density-based model & LOF   & ~\cite{ADG13,ADG36,ADG49,ADG53} \\
%     Energy-based model & DSEBM & ~\cite{ADG47,ADG48,ADG61,ADG73} \\
%     Autoencoder-based & DAGMM & ~\cite{ADG47,ADG48,ADG61,ADG73} \\
%     \bottomrule
%     \end{tabular}%
%   \label{tab:unsupervisedtechs}%
% \end{table}%

% Table generated by Excel2LaTeX from sheet 'Papers'
\begin{table}[tbp]
\renewcommand\thetable{17}
  \centering
 \small
  \caption{Unsupervised anomaly detection techniques compared to GANs.}
    \begin{tabular}{ll}
    \toprule
    \textbf{Techniques} & \textbf{List of references} \\
    \midrule
    Support vector machines & ~\cite{ADG03,ADG05,ADG30,ADG34,ADG36,ADG47,ADG48,ADG49,ADG52,ADG53,ADG60,ADG61,ADG71,ADG73,ADG99,ADG101,ADG108,ADG110,ADG115,ADG127} \\
    Ensemble methods & ~\cite{ADG03,ADG05,ADG34,ADG47,ADG49,ADG60,ADG61,ADG109,ADG110,ADG122} \\
    Principal component analysis & ~\cite{ADG11,ADG14,ADG22,ADG35,ADG37,ADG51,ADG62,ADG63,ADG64,ADG78,ADG79,ADG81,ADG85,ADG102,ADG108,ADG109,ADG120} \\
    Linear model & ~\cite{ADG14,ADG62,ADG85,ADG120} \\
    Probabilistic model & ~\cite{ADG20,ADG36,ADG53,ADG65,ADG110} \\
    Stochastic model & ~\cite{ADG62,ADG85,ADG120} \\
    Density-based model & ~\cite{ADG13,ADG36,ADG49,ADG53} \\
    Energy-based model & ~\cite{ADG47,ADG48,ADG61,ADG73} \\
    Autoencoder-based & ~\cite{ADG47,ADG48,ADG61,ADG73} \\
    \bottomrule
    \end{tabular}%
  \label{tab:unsupervisedtechs}%
\end{table}%

% \begin{figure}[htbp]
% \centering
% \includegraphics[width=1\linewidth]{PerformanceMetrics.pdf}
% \caption{Distribution of different performance metrics for anomaly detection.}
% \label{fig:PerformanceMetrics}
% \end{figure}

\section{Future Research Directions}\label{sec:futurework}

Generative adversarial network-based anomaly detection is in its early stage of development with many research opportunities. However, most of these opportunities lie in the field of GANs itself. In this section, we present possible directions for the future work of applying GANs in anomaly detection.

\textbf{Future direction 1: Speeding up the GAN training process.} Training GANs is a computationally demanding task. As reported in almost all primary studies, it takes a long time and powerful GPUs to train GANs to the point of satisfactory performance. Consequently, future studies need to explore GAN architectures that are lightweight and efficient in terms of resource consumption~\cite{ADG11,ADG22,ADG84,ADG101,ADG94}. For instance, the effects of selecting GAN hyperparameters on the anomaly detection performance have not been studied in the literature. There is also a need to consider the use of emerging GAN optimization and training methods, e.g.~\cite{arjovsky2017towards,salimans2016improved}, for better training stability and faster convergence.

\textbf{Future direction 2: Accounting for changing behaviour of a system.} In most industrial anomaly detection applications, behaviour of the target system varies over time. Therefore, it is crucial to examine the temporal behaviour of the system to find anomalies. RQ4 showed that only 7\% of the primary studies used GANs for anomaly detection in time series data. Therefore, more studies are required on anomaly detection using GANs for time series data to make them suitable for industrial applications, especially for multivariate time series data. 
In many industrial applications, data is collected online. Huang et al.~\cite{ADG01} suggest to take advantage of this data via online training of GAN-based anomaly detection techniques. This approach might be  adaptable for more real-time anomaly detection tasks.

\textbf{Future direction 3: Improving support for multimodal, discrete and noisy data.} Another open challenge of using GANs for anomaly detection is the lack of studies on multimodal anomaly detection using GANs. In real-world cases, data often comes from multiple sources with different types. For instance, Qiu et al.~\cite{ADG12} propose a GAN-based driving anomaly detection technique using physiological and CAN-bus data. Qiu et al. suggest incorporating other information such as results obtained from vision-based object detection systems applied to the road. Many other GAN-based anomaly detection approaches could benefit from using multimodal data. 
In addition, GANs were initially created to generate continuous data. As a result, they have limited ability to deal with discrete data, as it hinders the backpropagation process~\cite{ADG68}. Ben Fadhel and Nyarko~\cite{ADG68} point out this problem and study GAN architectures suitable for discrete data. Despite the promising results they report, this study is the only example of anomaly detection for discrete data in our review.
Finally, Lei et al.~\cite{ADG11} used optical flow as foreground shape information for video anomaly detection. Lei et al. point out that when the optical flow is inaccurate, it will affect the robustness of their proposed GAN-based anomaly detection technique. Thus, a potential direction for future research is the study of the effects of measurement inaccuracies and noise in the data on the performance of GAN-based anomaly detection techniques and the development of solutions to alleviate their impact.

%One of the biggest challenges in anomaly detection is that the normal behaviour of the system may change over time. An example of this problem is the change in the normal operating point of an industrial plant. We did not find any primary study investigating the problem of changes in the system's operating point or changes in the definition of normal behaviour.

\textbf{Future direction 4: Searching for better anomaly scoring methods.} As mentioned earlier, GAN-based anomaly detection techniques require an anomaly scoring method to distinguish between normal and abnormal samples. The selection of anomaly metrics for anomaly scores is still a challenging task. Further investigation is needed to improve the scoring methods and to identify the best fit for each application domain or for a specific case~\cite{ADG22,ADG62,ADG116}.

\textbf{Future direction 5: Improving the performance evaluation of GANs.} It is essential to evaluate the performance of GANs in generating data before using them for anomaly detection, either in a GAN-assisted or GAN-based setup. By doing so, one can ensure that GAN has learned the distribution of the data correctly. The results of RQ5 revealed that most primary studies do not evaluate the performance of their GAN-generated data. Additionally, almost all cases that assess data performance use image data. For other types of data, such as tabular, text and time series, there is no performance indicator of the generated data quality. Therefore, additional research is needed to identify the most suitable metrics for assessing the performance of GANs for each data type.

\textbf{Future direction 6: Employing improved GAN architectures for anomaly detection.} We observed in RQ3 that the `older' GAN architectures are by far the most popular in anomaly detection studies. However, many improved GAN architectures were proposed recently, which could improve anomaly detection as well.
For example, it is desirable to generate high-resolution images with GANs. However, it is a challenging task. High-resolution images make it easier for the discriminator to tell apart the generated images from the training samples~\cite{ACGANOdena}. 
Also, high-resolution images require more memory storage, which leads to smaller minibatches and may compromise training stability~\cite{PGGANkarras}. Several primary studies highlighted the need to generate high-resolution images for better anomaly detection, e.g.~\cite{ADG52,ADG66,ADG101}. Future studies may examine the effect of using improved GAN architectures, e.g., to improve the resolution of images using SRGAN~\cite{ledig2017photo}, ESRGAN~\cite{wang2018esrgan}, or BigGAN~\cite{brock2018large}, on the performance of anomaly detection techniques.
%Besides, this phenomenon could also be studied on time series data to increase the sampling resolution and possibly improve the anomaly detection accuracy. 

\section{Threats to Validity\label{sec:Threats}}
One threat to the validity of our review is that of missing papers.  The source of this threat is selecting the search keywords and the search engines. To mitigate this threat, we iteratively added keywords to our search query until no relevant new papers were found. The list of papers was finalized on the 3rd of June, 2020, and no papers published afterwards were added. It is possible that there are new GAN-based anomaly detection techniques that address some of the issues or challenges identified in this review.
%
%Another threat to the validity is that we did not carry out the quality assessment step for the systematic literature review. The reason for the decision is the subjective nature of the quality assessment. Therefore, all papers that satisfy the inclusion criteria have been included in the review process. 
%
Also, the data synthesis of the RQs was divided between the two first authors. To reduce bias in the data synthesis step, the first two authors met regularly to address disagreements. If a disagreement could not be resolved, one of the last two authors made the final decision.

\section{Conclusion}\label{sec:Conclusion}
This systematic literature review presents an extensive study on the applications of GANs for anomaly detection, covering 128 primary studies.  We define and answer several RQs to capture the current best practices and available techniques to employ GANs for anomaly detection purposes. We also identify the existing challenges and provide six future research directions in this area.

The results reveal that GANs are used for GAN-assisted (data augmentation) and GAN-based (representation learning) anomaly detection. In both cases,  the problem of insufficient amount of data for the anomalous behaviour of the system is addressed. In a GAN-assisted approach, the goal is to augment the minority class using the generative ability of GANs. In GAN-based anomaly detection, the goal is to use the representation learning ability of GANs, eliminating the need for minority class data. The most commonly used GAN architectures in the primary studies are DCGANs, standard GANs, and cGANs.
GANs are applied for anomaly detection in a wide range of application domains. The primary areas in which GANs are used for anomaly detection are medicine, surveillance and intrusion detection. However, their application in many other domains, such as anomaly detection in sensor networks, smart grids, and cloud computing shows that GANs are a suitable solution for anomaly detection. 

%We found that 50\% of the proposed GAN-based/GAN-assisted anomaly detection techniques are evaluated on image datasets. This confirms the ability of GANs to handle high-dimensional data. 

%The most commonly used GAN architectures in the primary studies are DCGANs, standard GANs, and cGANs. Despite their satisfactory generative capabilities, these architectures suffer from some problems, such as mode collapse and training instability. 
We identified six important directions for future research. Some of these directions are related to fundamental GAN research. For example, our study reveals that only 21\% of the primary studies evaluated the quality of the data that was generated with GANs. Hence, an important direction for future research is to investigate how the performance of GANs can be evaluated, as better performing GANs will also result in better performing anomaly detection approaches. Another important future research direction is speeding up the training process of GANs. %further evaluation of the generated data is important to assess whether it is a good representation of the learned distribution or not.
In addition, we identified several important future research directions for anomaly detection researchers. In particular, GAN-assisted anomaly detection approaches should improve their support for multimodal, discrete and noisy data, and account for changing behaviour of a system. Finally, researchers should investigate how recent improvements to GAN architectures can help improve their role in anomaly detection.

%GAN-based anomaly detection is usually performed in a semi-supervised manner. The cases described in primary studies used data from normal classes and trained their GAN to learn the normal data distribution. A score is assigned to the new data by a score function, and the anomalous data in the test stage is identified based on a  specific threshold. DCGANs and Standard GANs are the most popular architectures for semi-supervised anomaly detection using GAN. 

%In supervised learning-based anomaly detection, GANs are used to augment the dataset for the minority class. However, the primary studies report only a minor improvement in the anomaly detection technique's performance after augmenting the dataset with GANs. A few primary studies focused on pure unsupervised anomaly detection based on GANs, and most of them use the standard version of the GANs.

\section*{Acknowledgments}

The research reported in this article has been supported by the Government of Alberta under the Major Innovation Fund project RCP-19-001-MIF, and by the Natural Sciences and Engineering Research Council of Canada under the Alliance Grant project ALLRP 549804-19.

\bibliographystyle{unsrt}  
\bibliography{references}


\end{document}